\documentclass[10pt]{article} %
\usepackage[preprint]{tmlr}

\usepackage{amsmath,amsfonts,bm}

\def\eqref#1{equation~\ref{#1}}
\def\Eqref#1{Equation~\ref{#1}}

\def\1{\bm{1}}

\DeclareMathAlphabet{\mathsfit}{\encodingdefault}{\sfdefault}{m}{sl}
\SetMathAlphabet{\mathsfit}{bold}{\encodingdefault}{\sfdefault}{bx}{n}

\usepackage{hyperref}
\usepackage{svg}

\usepackage{url}
\usepackage{amsmath,amssymb}
\usepackage{amsthm}
\usepackage{graphicx}
\usepackage{subcaption}
\usepackage{algorithm} 
\usepackage{algpseudocode}
\usepackage{xcolor} 
\usepackage{booktabs}

\usepackage{tikz}
\usetikzlibrary{arrows.meta,calc}
\definecolor{tabcA}{HTML}{E69F00}
\definecolor{tabcB}{HTML}{56B4E9}
\definecolor{tabcC}{HTML}{009E73}
\definecolor{tabcD}{HTML}{CC79A7}

\theoremstyle{plain}           %
\newtheorem{theorem}{Theorem}

\newtheorem{remark}{Remark}
\newcommand{\cmp}[1]{\makebox[8.5em][l]{\text{#1}}} 
\newcommand{\grp}[1]{\makebox[8.5em][r]{\text{#1}}} 

\newtheorem{corollary}[theorem]{Corollary}
\newtheorem{lemma}[theorem]{Lemma}

\newtheorem{definition}{Definition}

\title{Temporally Abstract Bisimulation in Learned World Models}

\author{\name William Huang \email william.huang@yale.edu \\
      \addr Department of Computer Science\\
      Yale University
      \AND
      \name Benjamin Freed \email bfreed@andrew.cmu.edu \\
      \addr Robotics Institute\\
      Carnegie Mellon University
      \AND
      \name Howie Choset \email choset@andrew.cmu.edu\\
      \addr Robotics Institute\\
      Carnegie Mellon University}

\def\month{MM}  %
\def\year{YYYY} %
\def\openreview{\url{https://openreview.net/forum?id=XXXX}} %

\begin{document}

\maketitle

\begin{abstract}

    Agents capable of building temporally abstract representations of their environment through skills have been shown to perform better when planning over extended time horizons. Similarly, agents that filter high-dimensional observations into control-relevant state abstractions can plan more efficiently by discarding distracting information. Existing methods have largely explored both forms of abstraction in isolation. We discover that there exists a synergistic relationship between the two: when an agent plans over skills, a more aggressive state abstraction is possible because the skills can absorb lower-level control details so the planner's state does not need to encode them. We formalize this relationship with a temporally abstract bisimulation (TAB), a bisimulation on the skill-induced semi-MDP. Under this framework, we prove that temporal abstraction and state abstraction are synergistic, where composing skills can reduce the information required in the abstract state. Furthermore, we introduce a variational objective whose optima induce a TAB while jointly learning abstract states, skills, and a temporally and state-abstracted world model. On offline navigation benchmarks from D4RL, we show that the information the planner's abstract state must carry decreases as skill length is increased because low-level body state details are discarded. On OGBench navigation and manipulation benchmarks, our method plans directly over the learned abstract states and skills using sparse rewards without a value function or a graph. It achieves the second-best average rank against six model-free baselines and outperforms the graph-free planner of a recent model-based method in all environments.

\end{abstract}

\section{Introduction}

The human ability to develop \textit{models} of the world from past experience is a leading reason why humans are capable of efficiently completing new tasks \citep{Lenzen1944TheNO}. A model is a system that allows an intelligent agent, human or otherwise, to predict the effect of making a particular decision in a particular scenario. Models allow agents to \textit{plan} -- that is, evaluate many counterfactual decisions prior to executing them in the environment. This allows agents to make better use of past experiences, thereby allowing agents to learn more quickly and efficiently. Thus, the ability to plan is a cornerstone in intelligent decision-making in both AI and human psychology.

Planning becomes difficult in long-horizon and information-dense environments. There are a myriad of reasons for this, such as the curse of dimensionality \citep{bellman1957dynamic} and the credit assignment problem \citep{sutton1984temporal}. When we as humans plan, we primarily focus on the most task-relevant details, allowing us to better manage our limited cognitive resources \citep{ho2022people}. This is the process of \emph{abstraction}, which describes the ability to strip away low-level details to create a simplified model of a problem. Take the example of a driver who navigates to a grocery store. A simple plan might look something like this: make two right turns, one left turn, and then park in the lot. Notice how rather than planning over low-level actions like "drive forward 1 meter" the driver may instead break the task down into skills ("drive to the intersection" or "make a right turn") where the skill handles the low-level execution details. This process of reasoning over coarser units of time describes \textit{temporal abstraction}. Additionally, the states (parking lot, intersection, etc.) encountered at the termination of a skill are also abstracted, which describes \textit{state abstraction}.

Learning a decision-maker capable of reasoning with both temporal and state abstraction is an ongoing research problem in AI and robotics. Model-based deep reinforcement learning (MBRL) is a natural approach to learning such a decision-maker. In the previous driver and grocery store example, the driver's plan is itself a plan made with a model. The driver can compare the outcome of turning right against turning left at a particular intersection. A model-based learner should be able to learn and plan with a similar model, with skills as actions and abstract states as states. Additionally, model-based methods also promise high data-efficiency by reusing learned dynamics and can support transfer of knowledge across tasks.

Yet, model-based methods, particularly those that make heavy use of planning, have largely failed to scale to complex, high-dimensional environments. Learned world models are typically used to make predictions for a short future horizon, where subsequent predictions are deferred to a value function \citep{hafner2023dreamerv3, hansen2024tdmpc2scalablerobustworld}. As such, long-horizon benchmarks like OGBench \citep{park2025ogbench} have been primarily dominated by model-free approaches. Our key insight is that, similar to human decision-making, 
there is a synergy between the two abstractions, where the level of temporal abstraction can impact the achievable \textit{coarseness} of the state abstraction. Thus, methods that only learn each abstraction in isolation and don't optimize the two abstractions jointly are unable to capture this relationship. 

Expanding on this intuition, when planning over primitive actions, many of the state components are relevant for fine-grained control, so little can be discarded. Effective skills, however, absorb that control by responding to disturbances without invoking the high-level planner, which reduces the state information required by the planner. Returning to the previous example,
if the driver had chosen to operate over skills with a longer time horizon (such as directly drive to the grocery store), the driver only needs to know roughly where they are in the city, because the skill takes them there regardless of their current speed or lane. Conversely, a driver planning over shorter skills (such as make a left turn) must keep track of details such as current lane in the state because this information is key to whether the driver can successfully execute the skill.

In this work, we make this intuition precise through the lens of \emph{bisimulation}, which is a form of state abstraction that groups states together with the same rewards and transition probabilities \citep{hennessymilner, givan2003equivalence}. Because a bisimulation is defined relative to the dynamics, we can directly study how skills reshape the bisimulation relation. We define a temporally abstracted extension of bisimulation, called a \emph{temporally abstract bisimulation} (TAB), which characterizes when states can be considered behaviorally equivalent given their rewards and transition dynamics with skills. Unlike methods that combine the two abstractions as separate modules, we learn both from offline trajectories under a single variational objective whose reward and transition terms are exactly the two conditions of the bisimulation definition. We show theoretically that increasing temporal abstraction can yield a coarser state abstraction, and empirically that a planner can use a state representation with fewer bits of information when the temporal abstraction is increased. In summary, the contributions of this paper are as follows:
\begin{enumerate}
    \item We provide a theoretical motivation for learning joint temporal and state abstractions with a bisimulation defined over skills rather than primitive actions, which we formalize through the idea of a temporally abstract bisimulation.
    \item We derive a novel variational objective that
    jointly infers temporally extended skills and compact abstract states from
    offline trajectories, and prove that its optima induce a temporally abstract bisimulation.
    \item We show theoretically and empirically that there is a synergy between temporal and state abstraction, where increasing the coarseness of the temporal abstraction can enable a coarser state abstraction. Further, we demonstrate that our model, which learns neither a value function nor uses shaped rewards, is competitive with model-free and model-based baselines on navigation and manipulation benchmarks from OGBench.
\end{enumerate}

\section{Background} \label{sec:background}

In this section, we build on the theory of Markov Decision Processes (MDPs) and Partially Observable MDPs (POMDPs). An MDP is defined by a tuple $(\mathcal{S}, \mathcal{A}, P, R, \gamma)$ with a finite state space $\mathcal{S}$, action space $\mathcal{A}$, transition function $P(s
'\mid s, a)$ modeling the probability of transitioning from state $s \in \mathcal{S}$ to state $s' \in \mathcal{S}$, reward function $R(s, a)$, and discount $\gamma \in [0,1)$. At each timestep, the agent in state $s_t$ takes action $a_t \sim \pi(\cdot \mid s_t)$ under policy $\pi$ and receives reward $r_t=R(s_t, a_t)$. This definition assumes the agent observes $s_t$. When it does not, as in a POMDP, the agent instead receives observation $o_t \sim \Omega(\cdot \mid s_t)$ in a finite observation space $\mathcal{X}$. The agent then conditions on the history $h_t = (o_0, a_0, o_1, \dots, a_{t-1}, o_t)$ which is Markovian.

\subsection{Temporal Abstraction}
\label{sec:semimdp}
Temporal abstraction has been formalized mathematically using the \textit{options} framework \citep{sutton1999between}. An \emph{option} is an extended notion of an action represented by a closed-loop policy that takes actions over an extended period of time. Following \citet{sutton1999between}, formally, an option is $z = \langle \mathcal{I}_z,\ \pi_z,\ \beta_z \rangle$,
with an initiation set $\mathcal{I}_z \subseteq \mathcal{S}$, a policy
$\pi_z(a \mid s)$, and a termination condition $\beta_z(s) \in [0,1]$. Option $z$ is available at $s$ if and only if $s \in \mathcal{I}_z$. Once initiated, it selects actions according to $\pi_z$ and terminates with probability $\beta_z$. We write $\mathcal{Z}$ for the set of options. A set of options defined over an MDP induces a \emph{semi-MDP}. 

There has been extensive recent literature about learning options-based temporal abstractions. In modern deep learning, options are synonymous with \textit{skills} \citep{ eysenbach2018diversityneedlearningskills, Pertsch2020AcceleratingRL}. Skill learning can largely be categorized into model-free and model-based approaches. In model-free methods, the skills are either learned online, by maximizing the mutual information between skills and the trajectories they produce \citep{eysenbach2018diversityneedlearningskills, achiam2018variationaloptiondiscoveryalgorithms}, or offline, by encoding segments of demonstration data into skills with a latent variable model \citep{Pertsch2020AcceleratingRL, ajay2020opal, pmlr-v119-shankar20b}. However, model-free approaches do not learn how skills influence the state of the environment, so they cannot plan. Model-based approaches instead learn a model over the dynamics of skill transitions; these methods largely approach this problem from a variational inference perspective \citep{freed2023oposm, sharma2020dynamicsawareunsuperviseddiscoveryskills}. Here, a skill-conditioned low-level policy is learned to map state-skill pairs to actions.

\subsection{State Abstraction} \label{sec:bisim}
The central idea behind state abstraction is to reduce the size of the state space by grouping states together into \emph{abstract states} in a way that does not change anything important about the underlying problem. State abstraction approaches primarily differ in what information is treated as important and thus should be preserved. At one end, the abstract state is encouraged to contain sufficient information for reconstruction. These methods train an encoder  to map each observation to a latent state, from which the original observation and reward is decoded \citep{hafner2019planet, hafner2020dreamer, hafner2023dreamerv3}. 
At the other end, the abstract state groups observations into clusters such that information relevant for optimal control is maintained. For example, $Q^*$-irrelevance groups states that have the same optimal action values, and $\pi^*$-irrelevance groups states that share an optimal action \citep{li2006towards}. A similar constraint called the value equivalence principle enforces that a learned model only needs to produce the same Bellman updates as the environment for a chosen set of policies and value functions \citep{grimm2020value}. In this paper, we focus on one particular mathematical framework for state abstraction: bisimulation.

\paragraph{State Abstraction with Bisimulation}
Bisimulation is one of the strictest forms of state abstraction that
preserves key properties such as relevant dynamics and optimal value  \citep{Larsen1991BisimulationTP}.
For an MDP $(\mathcal{S}, \mathcal{A}, P, R, \gamma)$, a relation $\sim$ on $\mathcal{S}$ is a \emph{bisimulation} if two bisimilar states share immediate rewards and transition into states that are also bisimilar. Formally,
if $s_i \sim s_j$,
\begin{align}
R(s_i, a) &= R(s_j, a) \quad \forall a \in \mathcal{A}, \label{eq:bisim-rew}\\
P(C \mid s_i, a) &= P(C \mid s_j, a) \quad \forall a \in \mathcal{A},\ \forall C \in \mathcal{S}/\!\sim \label{eq:bisim-dyn}
\end{align}

where $P(C \mid s, a) = \sum_{s' \in C} P(s' \mid s, a)$. The coarsest such relation partitions $\mathcal{S}$ into \emph{bisimulation equivalence classes}. These classes can be treated as abstract states that are behaviorally equivalent \emph{without loss of optimal control} \citep{givan2003equivalence, li2006towards}. Therefore, bisimulation is a definitive framework for state abstraction. However, this definition assumes the agent observes the Markov state $s_t$. When the agent does not, as in a POMDP, the history is Markovian. We can then think of bisimulation relations as partitioning histories rather than states. We work with histories throughout this work.

In deep RL, bisimulation-based state abstractions are learned by matching bisimulation metrics \citep{zhang2021learning} or by learning exact bisimulation relations \citep{vibes}. However, these works learn the state abstraction with respect to primitive actions and do not consider a temporally extended bisimulation defined with respect to a set of skills.

\subsection{Joint State and Temporal Abstraction}
State and temporal abstraction have also been studied together.  \textit{SMDP homomorphisms} are maps on state--action pairs, where actions can be temporally extended, that preserve rewards and abstract transition dynamics at the level of options \citep{Ravindran2003, Ravindran2003b}. Other works derive value-preserving state abstractions for a given set of options \citep{Abel2020}. Yet, these methods assume the options are fixed and derive the state abstraction from them. In deep RL, hierarchical methods learn skills together with a latent model. Director, for instance, learns a framework in which a high-level
manager proposes latent subgoals that a low-level worker realizes for $K$ steps \cite{hafner2022deephierarchicalplanningpixels}. However, the learned latent state is independent of $K$ because the world model only makes predictions one-step ahead and is trained independently of the hierarchy. In contrast, we learn skills, abstract states, and a temporally and state-abstracted world model through an objective whose optima are bisimulations defined with respect to a skill set. Because the bisimulation depends on the skills, learning them jointly allows us to study the \emph{synergy} between the two abstractions.

\begin{figure}[t]
  \centering
  \includegraphics[scale=0.5]{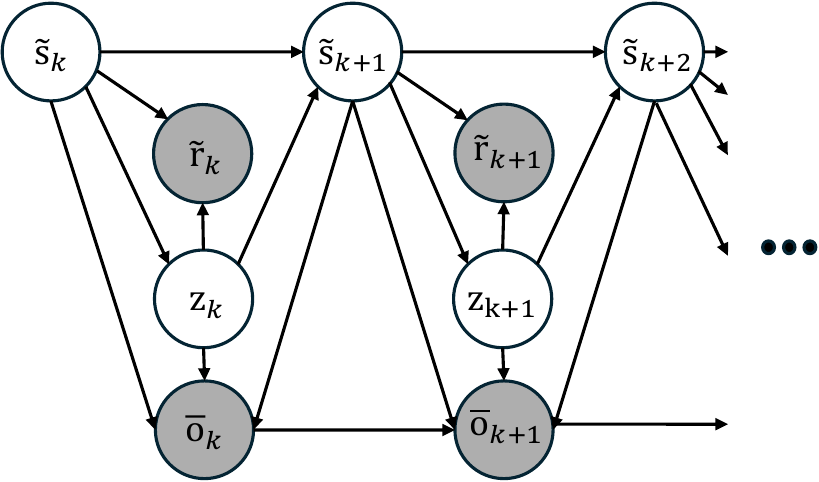}
    \caption{\textbf{Graphical model of the temporally and state-abstracted world model.}
    Latent variables (unshaded) are the abstract boundary states $\tilde{s}_k$ and skills $z_k$. Observed variables (shaded) are the segment observations/actions $\bar{o}_k$, $\bar{a}_k$ and the segment reward $\tilde{r}_k$.}
  \label{fig:causalgraph}
\end{figure}

\section{Temporally Abstract Bisimulation Relations}
\label{sec:TAB}

In this section, we propose a formalization of joint state and temporal abstraction through a \emph{temporally abstract bisimulation (TAB)}. A temporally abstract bisimulation is a bisimulation on the semi-MDP induced by applying option-based temporal abstraction. We show that, like how a primitive bisimulation enables optimal control on the original MDP, TABs enable optimal control on the semi-MDP we apply bisimulation to. Moreover, we prove temporal abstraction and state abstraction are synergistic: the more options are composed, the less information the state may need to encode.

\subsection{Temporally Abstract Bisimulation} \label{sec:tab}
The bisimulation described in \S\ref{sec:bisim} is defined over primitive
transitions. To combine temporal and state abstraction, we extend the primitive bisimulation to the option-induced semi-MDP.

A trajectory is segmented at boundaries $t_0 = 0$, $t_{k+1} = t_k + \tau_k$, where $\tau_k$ is the actual duration of the $k$-th option $z_k$.
We write $h_k = h_{t_k}$ for the history at boundary $k$. We write $\bar{o}_k = (o_{t_{k}}, \dots, o_{t_{k+1}})$ and
$\bar{a}_k = (a_{t_k}, \dots, a_{t_{k+1}-1})$ for the observations and actions within
segment $k$, and define the
segment reward as $\tilde{r}_k = \sum_{i=0}^{\tau_k - 1} \gamma^{i}\,R\big(s_{t_k + i},\, a_{t_k + i}\big)$.  Executing option $z_k$ from boundary $k$ gives a segment reward $\tilde{r}_k$. This brings the agent to the next boundary state, where the history is
$h_{k+1}$.

\begin{definition}[Temporally Abstract Bisimulation Relations]
\label{def:bisim}
Given a POMDP and a set of options $\mathcal{Z}$, let $\mathcal{Z}(h_k) = \{z \in \mathcal{Z} : o_{t_k} \in \mathcal{I}_z\}$ denote the options available at $h_k$. An equivalence relation is a temporally abstract bisimulation relation if, for every $k$, the class of $h_k$ determines which options are available and is a sufficient statistic of the history for the discounted transition between abstract states and for the expected segment reward. That is, for all $h_k \sim h_k'$,
\begin{align}
\text{(availability)} \quad & \mathcal{Z}(h_k) = \mathcal{Z}(h_k'), \label{eq:avail}\\
\intertext{and for every skill $z_k \in \mathcal{Z}(h_k)$,}
\text{(dynamics sufficiency)} \quad & K_\gamma(C \mid h_k, z_k) = K_\gamma(C \mid h_k', z_k) \quad \text{for every class } C \text{ of } \sim, \label{eq:dyn-suff-gen}\\
\text{(reward sufficiency)}   \quad & \mathbb{E}[\tilde{r}_k \mid h_k, z_k] = \mathbb{E}[\tilde{r}_k \mid h_k', z_k], \label{eq:rew-suff-gen}
\end{align}
where $K_\gamma(C \mid h_k, z_k) = \mathbb{E}\big[\gamma^{\tau_k}\, \mathbf{1}\{h_{k+1} \in C\} \,\big|\, h_k, z_k\big]$ is the discounted probability that option $z_k$ terminates in class $C$. We write $\mathcal{Z}_{C}$ for the options available at the histories in class $C$.
\end{definition}

This dynamics sufficiency condition plays the role of \eqref{eq:bisim-dyn} of a primitive bisimulation. However, because options can take a variable number of steps, it weights each transition by $\gamma^{\tau_k}$. If options have fixed-length $H$, then $\tau_k = H$ and $K_\gamma(C \mid h_k, z_k) = \gamma^H P(h_{k+1} \in C \mid h_k, z_k)$. If $\gamma=1$, $K_1(C \mid h_k, z_k) = P(h_{k+1} \in C \mid h_k, z_k)$ for options of any length. In both cases, when $\gamma > 0$
, these equations match the condition of \eqref{eq:bisim-dyn} except with actions replaced by options. Availability holds trivially when every option is available at every observation.

\begin{theorem}[TAB supports optimal control in the option-induced semi-MDP]
\label{thm:optimal-control}

Let $e$ be a temporally abstract bisimulation, and let $\mathcal{M}$ denote the semi-MDP over histories. Then any optimal policy of $\mathcal{M}/\!\sim_e$ induces an optimal policy of $\mathcal{M}$ that selects options from the class of the current history.
\end{theorem}
\begin{proof}[Proof Sketch]
Because equivalent histories have the same available skills, the same expected segment reward, and the same discounted transition into every class, we can form the quotient $\mathcal{M}/\!\sim$, a semi-MDP whose states are the classes of $\sim$. We show by value iteration that each history has the same optimal value as its class, so the optimal skill values also agree. As in Theorem 7 of \citet{givan2003equivalence}, an optimal policy of $\mathcal{M}/\!\sim$ then selects an optimal skill at every history. The full proof is in Appendix~\ref{app:optimal}.
\end{proof}

\begin{theorem}
[Existence of the coarsest temporally abstract bisimulation]
\label{thm:coarsest}
For a fixed skill set $\mathcal{Z}$, there is a unique coarsest temporally abstract bisimulation $\sim_{\mathrm{TA}}$. Any two histories that are related by some temporally abstract bisimulation for $\mathcal{Z}$ are related by $\sim_{\mathrm{TA}}$.  Full proof is in Appendix~\ref{app:coarsest}.
\end{theorem}

\begin{figure}[t]
\centering
\includegraphics[width=0.8\linewidth]{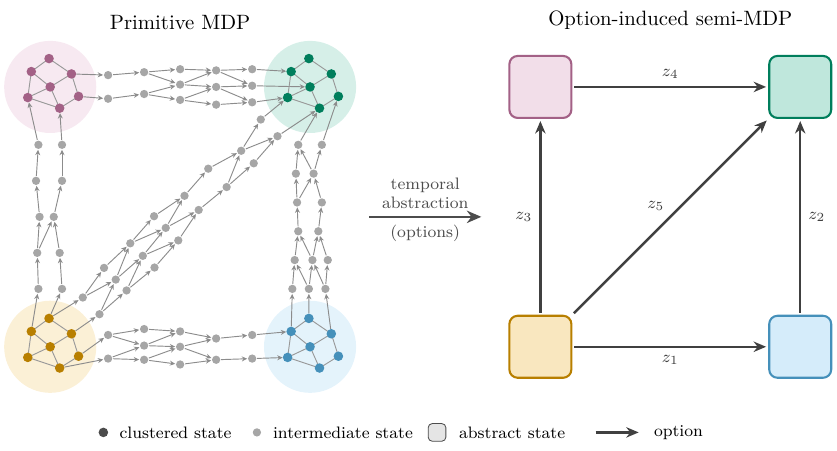}
\caption{Intuition behind why temporal abstractions can admit a coarser state abstraction. On the left, the primitive bisimulation must keep states in each cluster distinct because their one-step transitions and rewards are distinct. In contrast, if the states in each cluster have the same discounted transitions and expected segment rewards under each option, each cluster can be grouped into a single abstract state as seen on the right. Moreover, the intermediate states do not belong to any option initiation set and are abstracted away by the options.}
\label{fig:temporal-coarsening}
\end{figure}

\subsection{Coarser Temporal Abstractions Can Enable Coarser State Abstractions}
We show that coarsening the temporal abstraction should also coarsen the state abstraction. Intuitively, a TAB only needs to distinguish histories when an option leads them to abstract states with a different discounted probability and different expected segment rewards; with longer skills, predicting expected rewards and discounted transitions between skill boundaries is no longer relevant because they are absorbed by the skill, so fewer distinctions should be required. We depict this intuition in Figure~\ref{fig:temporal-coarsening}. A skill composition is a special form of temporal coarsening because a composed skill is at least as long as the shorter skills it combines. We prove rigorously for composition that coarsening the temporal abstraction permits a coarser state abstraction.   

\begin{remark}[Primitive bisimulation is a TAB]
\label{rem:prim-special-case}
A primitive bisimulation is a temporally abstract bisimulation when the options are the primitive actions. Suppose there is one option for each $a \in \mathcal{A}$ that takes action $a$ once and terminates, and each is available at every observation, so \eqref{eq:avail} holds trivially. Then $\tau_k = 1$, $K_\gamma(C \mid h_t, z_k) = \gamma P(h_{t+1} \in C \mid h_t, a)$, and $\mathbb{E}[\tilde{r}_k \mid h_t, z_k] = \mathbb{E}[R(s_t, a) \mid h_t]$. These are the transition and reward of the MDP over histories, so for $\gamma > 0$, \eqref{eq:dyn-suff-gen} and \eqref{eq:rew-suff-gen} are \eqref{eq:bisim-dyn} and \eqref{eq:bisim-rew} for that MDP, and every primitive bisimulation on histories is a temporally abstract bisimulation.
\end{remark}

It then directly follows that, over all choices of possible sets of options, the coarsest TAB is at least as coarse as the primitive bisimulation because the primitive bisimulation is the TAB of the semi-MDP induced by the option set where each option is an $H=1$ primitive action.

\begin{definition}[Composed options]
\label{def:composition}
Let $z_1, z_2 \in \mathcal{Z}$ and suppose $z_2$ is available at every observation at which $z_1$ can terminate. The \emph{composition} $z_1 z_2$ is the option that executes $z_1$ until termination and then executes $z_2$ until termination. Write $\mathcal{Z}^{(m)}$ to denote the set of $m$-fold compositions $z_1 z_2 \cdots z_m$ of options in $\mathcal{Z}$, so $\mathcal{Z}^{(1)} = \mathcal{Z}$.
\end{definition}

\begin{theorem}[Composition preserves temporally abstract bisimulation]
\label{thm:composition}
Let $\sim$ be a temporally abstract bisimulation for the option set $\mathcal{Z}$. Then $\sim$ is a temporally abstract bisimulation for $\mathcal{Z}^{(m)}$, for every $m \ge 1$.
\end{theorem}

\begin{proof}[Proof sketch]
We prove by induction on $m$ that, for every composed option in $\mathcal{Z}^{(m)}$, the option availability, the expected segment reward, and the discounted transition of the next abstract state are dependent on the history only through the class from which the option is initiated. Any composition of $m+1$ options can be split into a composition $z_1$ of $m$ options and an additional option $z_2$. The composed option is available where the first option is available so by the induction hypothesis the availability of $z_1z_2$ is dependent on on the history only through the class. The discounted probability that $z_1z_2$ terminates is a sum over the class where $z_1$ terminates of the product between the discounted probability of reaching that class multiplied by the discounted probability of reaching the target class under $z_2$. The expected segment reward is the expected reward of $z_1$ plus the sum over the possible classes from which $z_1$ terminates of the discounted expected reward of $z_2$ beginning from those classes. By the induction hypothesis and the TAB conditions for $z_2$, every term depends on the history only through the starting class. Full proof is in Appendix~\ref{app:composition}.

\end{proof}
\begin{corollary}[Composed options admit coarser abstractions]
\label{cor:nesting}
Write $\sim_{\mathrm{TA}}(\mathcal{Z})$ for the coarsest temporally abstract bisimulation with respect to the option set $\mathcal{Z}$. Then $\sim_{\mathrm{TA}}(\mathcal{Z}) \subseteq \sim_{\mathrm{TA}}(\mathcal{Z}^{(m)})$ for every $m \ge 1$. In particular, along a chain in which each option length divides the next, such as $\mathcal{Z}, \mathcal{Z}^{(2)}, \mathcal{Z}^{(4)}, \ldots$, we have $\sim_{\mathrm{TA}}(\mathcal{Z}) \subseteq \sim_{\mathrm{TA}}(\mathcal{Z}^{(2)}) \subseteq \sim_{\mathrm{TA}}(\mathcal{Z}^{(4)}) \subseteq \cdots$.
\end{corollary}

\begin{proof}
By Theorem~\ref{thm:composition}, $\sim_{\mathrm{TA}}(\mathcal{Z})$ is a temporally abstract bisimulation for $\mathcal{Z}^{(m)}$. By Theorem~\ref{thm:coarsest}, $\sim_{\mathrm{TA}}(\mathcal{Z}^{(m)})$ contains every temporally abstract bisimulation for $\mathcal{Z}^{(m)}$, so it also contains $\sim_{\mathrm{TA}}(\mathcal{Z})$. Along the chain, apply the same argument with $\mathcal{Z}^{(m)}$ as the skill set. Since $(\mathcal{Z}^{(m)})^{(n)} = \mathcal{Z}^{(mn)}$, this gives $\sim_{\mathrm{TA}}(\mathcal{Z}^{(m)}) \subseteq \sim_{\mathrm{TA}}(\mathcal{Z}^{(mn)})$ for every $n \ge 1$.
\end{proof}

\begin{figure*}[t]   %
  \centering
  \begin{minipage}[c]{0.40\textwidth}
    \begin{algorithm}[H]
    \caption{Learning a Temporally and State-Abstracted World Model for Zero-Shot Planning}
    \label{alg:salt}
    \small
    \begin{algorithmic}
    \State \textbf{Training Input}: Dataset $\mathcal{D}$, skill length $H$, number of skills $K$
    \State \textcolor{gray}{=== Training Phase ===}
    \Repeat
      \State $x = (o_{0:KH}, a_{0:KH-1}, g) \sim \mathcal{D}$
      \State $\tilde{s}_k \sim q_\chi(\tilde{s}_k \mid h_k)$, \ $k = 0, \dots, K$
      \State $z_k \sim q_\phi(z_k \mid \bar{o}_k, \bar{a}_k, \tilde{s}_k)$, \ $k = 0, \dots, K-1$
      \State gradient step on $\mathcal{L}(x)$
    \Until{converged}
    \State \textcolor{gray}{=== Decision-Time Planning ===}
    \State \textbf{Testing Input}: $o_0$, $g$, planning horizon $L$
    \State $h \gets (o_0)$
    \While{goal not reached}
      \State $\tilde{s} \gets \mathbb{E}_{q_\chi}[\tilde{s} \mid h]$
      \State $\epsilon^\star_{1:L} \gets \mathrm{CEM}(\tilde{s}, g, p_\omega, p_\psi, p_\rho)$
      \State $z \gets \mu_\omega(\tilde{s}) + \sigma_\omega(\tilde{s}) \odot \epsilon^\star_1$
      \For{$t = 1, \dots, H$}
        \State $a \sim \pi_\theta(a \mid o, z)$
        \State $o \gets \text{env.step}(a)$, \ $h \gets (h, a, o)$
      \EndFor
    \EndWhile
    \end{algorithmic}
    \end{algorithm}
  \end{minipage}\hfill
  \begin{minipage}[c]{0.60\textwidth}
    \centering
    \includegraphics[width=\linewidth]{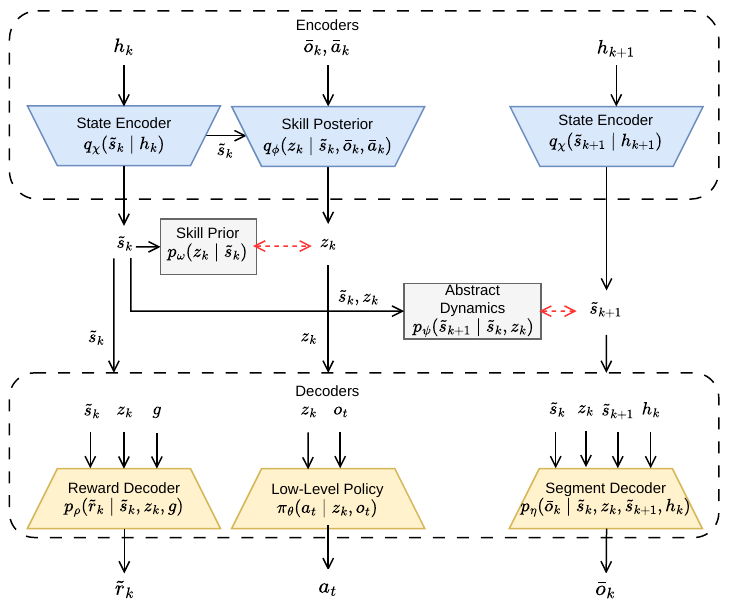}
    \caption{\textbf{Temporally and State-Abstracted Model Diagram and Objective}. We infer abstract boundary states and latent skills from offline trajectory segments, predict skill-conditioned transitions in the learned abstract state space, and score imagined transitions for goal-conditioned planning. Red dotted arrows denote the KL-matching terms of the ELBO~\eqref{eq:elbo}.}
    \label{fig:architecture}
  \end{minipage}
\end{figure*}

\section{Learning Temporally Abstract Bisimulations}
\label{sec:learning}

In \S\ref{sec:TAB} we characterized temporally abstract bisimulations without specifying how to learn one. In this section, we introduce an approach to learn a TAB by simultaneously learning temporally extended skills, abstract states, and a skill-conditioned temporally and state-abstracted world model (TSAWM) from an offline dataset. Together, these components enable zero-shot planning on a downstream task over the abstract skill and state space. The pseudocode for the algorithm training and planning is outlined in Algorithm \ref{alg:salt} with its components detailed in Figure \ref{fig:architecture}.

Here, we follow the convention from previous work in which skills are represented by a skill-conditioned low-level policy $\pi_\theta(a \mid o, z)$ where $a \in \mathcal{A}$ is the agent action, $o \in \mathcal{X}$ is the current raw observation, and $z \in \mathcal{Z}$ is a skill variable. The policy does not condition on the history because the environments used in our experiments are fully observable MDPs and thus the observation contains the full state. We consider skills of fixed length $H$ so $\tau_k=H$. Additionally, each skill can be initiated at every observation, since $\pi_\theta(a \mid o, z)$ conditions on any $z$ so $\mathcal{I}_z=\mathcal{X}$. Therefore, \eqref{eq:avail} holds automatically and $K_\gamma(C \mid h_k, z_k) = \gamma^H P(h_{k+1} \in C \mid h_k, z_k)$. 

A state encoder $e$ that maps a history $h_k$ to an abstract state $\tilde{s}_k$ induces an equivalence relation on histories, $h \sim_e h' \iff e(h) = e(h')$, whose classes are the level sets of $e$. The encoder $e$ represents a TAB when $\sim_e$ is one. 

Our model predicts the full distribution of the next abstract state and segment reward to follow the conditions
\begin{align}
\text{(dynamics sufficiency)} \quad & e(h_{k+1}) \;\perp \! \! \! \perp\; h_k \;\mid\; \tilde{s}_k,\, z_k, \label{eq:dyn-suff}\\
\text{(reward sufficiency)}   \quad & \tilde{r}_k \;\perp \! \! \! \perp\; h_k \;\mid\; \tilde{s}_k,\, z_k. \label{eq:rew-suff}
\end{align}
\Eqref{eq:rew-suff} is a stronger condition than \eqref{eq:rew-suff-gen} because it is a constraint on the entire distribution, not just the expectation. For the fixed-length skill case, since $\gamma > 0$, \eqref{eq:dyn-suff} is equivalent to \eqref{eq:dyn-suff-gen} because the classes of $\sim_e$ are the level sets of $e$. Therefore, an encoder that satisfies \eqref{eq:dyn-suff} and \eqref{eq:rew-suff} induces a TAB. The theory in the previous section also permits learning skills with variable durations, which can be done by also learning a skill termination condition. With fixed-length skills, a trajectory of length $T = KH$ splits into $K$ segments. As in related works, $T$ is a fixed parameter that is shorter than the total episode length \citep{ajay2020opal, pertsch2021spirl, eysenbach2018diversityneedlearningskills, achiam2018variationaloptiondiscoveryalgorithms}.

\subsection{Model Components} 
We adopt a variational inference view to learn (1) a skill prior over the skills available from an abstract state; (2) a reward decoder predicting segment return after executing a skill; (3) an abstract dynamics model predicting skill-conditioned abstract state transitions; (4) a segment decoder
reconstructing the observations of a segment from its boundary states and skill; (5) a low-level policy mapping skills and observations to primitive actions at each time step; (6) a state encoder inferring abstract state from the observation-action history and (7), a skill posterior inferring the current skill from segment-level observations and actions. Together, along with an initial abstract state prior and initial observation decoder, we posit the following components: 

\begin{equation}
\label{eq:components}
\begin{aligned}
&\grp{}\;\hphantom{\{\,}
\begin{array}{@{}l@{\quad}l@{}}
  \cmp{Initial state prior:}   & \tilde{s}_0 \sim p(\tilde{s}_0) \\
  \cmp{Initial decoder:}       & o_0 \sim p_{\eta_0}(o_0 \mid \tilde{s}_0) \\
  \cmp{Skill prior:}           & z_k \sim p_\omega(z_k \mid \tilde{s}_k)
\end{array} \\
&\grp{TAB Conditions}\;\begin{cases}
  \cmp{Abstract dynamics:}         & \tilde{s}_{k+1} \sim p_\psi(\tilde{s}_{k+1} \mid \tilde{s}_k, z_k) \\
  \cmp{Reward decoder:}           & \tilde{r}_k \sim p_\rho(\tilde{r}_k \mid \tilde{s}_k, z_k, g)
\end{cases} \\
&\grp{}\;\hphantom{\{\,}
\begin{array}{@{}l@{\quad}l@{}}
  \cmp{Segment decoder:}   & \bar{o}_k \sim p_\eta(\bar{o}_k \mid \tilde{s}_k, z_k, \tilde{s}_{k+1}, h_k) \\
  \cmp{Low-level policy:}      & a_t \sim \pi_\theta(a_t \mid o_t, z_k), \quad kH \le t < (k+1)H \\
  \cmp{State encoder:}         & \tilde{s}_k \sim q_\chi(\tilde{s}_k \mid h_k) \\
  \cmp{Skill posterior:}         & z_k \sim q_\phi(z_k \mid \bar{o}_k, \bar{a}_k, \tilde{s}_k)
\end{array}
\end{aligned}
\end{equation}

\paragraph{Goal-conditioned reward.}
The theory in Section~\ref{sec:tab} is stated for a single reward function. We train our model in goal-reaching tasks like OGBench from \S\ref{sec:exp-domains} where the reward depends on a goal $g$ while the dynamics do not, so each goal defines its own MDP with reward $R(\cdot, \cdot\,; g)$ and the same transitions. Our reward head takes the goal as an input such that it can represent the reward model for every goal. All other models do not observe $g$, and we suppress $g$ in the notation below. Theorem~\ref{thm:elbo-tab} then applies to each goal separately. The objective also averages over the goals in the data, and the proof is unchanged with $g$ as an additional conditioning variable. Our encoder does not condition on $g$. Thus, its optimum is a single abstraction that is a temporally abstract bisimulation for every goal in the set of possible training goals.

\subsection{Generative Model and Objective}
\label{sec:model}

Write $x = (o, a, \tilde{r})$ for an observed trajectory. The latent variables are the
abstract boundary states $\tilde{s}_0, \ldots, \tilde{s}_K$ and the skills $z_0, \ldots, z_{K-1}$
(Figure~\ref{fig:causalgraph}). The generative model is
\begin{align}
    p_\Theta(x, \tilde{s}, z)
      &= p(\tilde{s}_0)\, p_{\eta_0}(o_0 \mid \tilde{s}_0)
      \prod_{k=0}^{K-1}
         \Big[\, p_\omega(z_k \mid \tilde{s}_k)\,
                 p_\rho\big(\tilde{r}_k \mid \tilde{s}_k, z_k\big)\,
                 p_\psi\big(\tilde{s}_{k+1} \mid \tilde{s}_k, z_k\big) \notag \\
      &\qquad \times p_\eta(\bar{o}_k \mid \tilde{s}_k, z_k, \tilde{s}_{k+1}, h_k)
       \prod_{t = kH}^{(k+1)H - 1}
                \pi_\theta(a_t \mid o_t, z_k) \Big],
    \label{eq:generative}
\end{align}
and is parameterized by $\Theta = (\eta_0, \omega, \rho, \psi, \eta, \theta)$.

The abstract dynamics model $p_\psi$ and reward decoder $p_\rho$ correspond to dynamics sufficiency (\eqref{eq:dyn-suff}) and reward sufficiency (\eqref{eq:rew-suff}), respectively. The segment decoder $p_\eta$ is allowed to see the history $h_k$, which provides an alternative pathway besides the abstract states to obtain information about the segment observations. This removes the pressure from the abstract states to encode distracting information about the segment observations for reconstruction, instead freeing them to focus on features relevant for predicting segment rewards and abstract state transitions. When combined with the information bottleneck applied to the encoder (described below), the abstract states is encouraged towards a minimal representation, the coarsest temporally abstract bisimulation, keeping only the information required to satisfy the two sufficiency conditions. If the segment decoder did not condition on the history, the abstract state would be pushed to also encode distracting features relevant for reconstruction.

We infer the latents with the following posterior,
\begin{equation}
    q_\Phi(\tilde{s}, z \mid x) = q_\chi(\tilde{s}_0 \mid h_0) \prod_{k=0}^{K-1}
        q_\phi(z_k \mid \bar{o}_k, \bar{a}_k, \tilde{s}_k)\; q_\chi(\tilde{s}_{k+1} \mid h_{k+1}),
    \label{eq:posterior}
\end{equation}
with parameters $\Phi = (\phi, \chi)$. The encoder $q_\chi$ maps a history
to an abstract boundary state, exactly as $e$ does above. In practice, our encoder outputs a Gaussian with a fixed variance and a mean bounded to $[-M, M]$ in each coordinate, thus fixing an upper bound on the amount of information that can be encoded into the abstract state.

The objective is the evidence lower bound
\begin{align}
    \mathcal{L}(\Theta, \Phi; x)
      &= \mathbb{E}_{q_\Phi}\Big[\, \sum_{k=0}^{K-1} \Big(
         \log p_\eta(\bar{o}_k \mid \tilde{s}_k, z_k, \tilde{s}_{k+1}, h_k)
           + \sum_{t=kH}^{(k+1)H-1} \log \pi_\theta(a_t \mid o_t, z_k) \notag\\
      &\qquad\qquad
         + \log p_\rho\big(\tilde{r}_k \mid \tilde{s}_k, z_k\big)
         - \mathrm{KL}\big[\, q_\chi(\tilde{s}_{k+1} \mid h_{k+1}) \,\big\|\,
                p_\psi(\tilde{s}_{k+1} \mid \tilde{s}_k, z_k) \,\big] \notag\\
      &\qquad\qquad
         - \mathrm{KL}\big[\, q_\phi(z_k \mid \bar{o}_k, \bar{a}_k, \tilde{s}_k) \,\big\|\,
                p_\omega(\cdot \mid \tilde{s}_k) \,\big] \Big)
         + \log p_{\eta_0}(o_0 \mid \tilde{s}_0) \Big] - \mathrm{KL}\big[\, q_\chi(\tilde{s}_0 \mid h_0) \,\big\|\, p(\tilde{s}_0) \,\big],
    \label{eq:elbo}
\end{align}
which we maximize in expectation over the data,
$\mathcal{L}(\Theta, \Phi) = \mathbb{E}_{x \sim p_{\mathrm{data}}}\big[\mathcal{L}(\Theta, \Phi; x)\big]$. A full derivation is deferred to
Appendix~\ref{app:derivation}.

\subsection{Why $\mathcal{L}$ Yields a Temporally Abstract Bisimulation}
\label{sec:why}

We formalize the guarantee with a lemma (sufficiency $\Rightarrow$ temporally abstract bisimulation) and
a theorem (optimality of $\mathcal{L}$ $\Rightarrow$ sufficiency). Full proofs are in
Appendix~\ref{app:proofs-elbo}.

\begin{lemma}[Sufficiency induces a temporally abstract bisimulation]
\label{lem:suff-tab}
Let $\tilde{s}_k = e(h_k)$. Suppose $e$ satisfies dynamics sufficiency (\eqref{eq:dyn-suff}) and reward
sufficiency (\eqref{eq:rew-suff}). Then for all
$h_k \sim_e h_k'$ and all $z \in \mathcal{Z}$,
\begin{align*}
    P\big(e(h_{k+1}) \mid h_k, z\big) &= P\big(e(h_{k+1}) \mid h_k', z\big), \\
    P\big(\tilde{r}_k \mid h_k, z\big) &= P\big(\tilde{r}_k \mid h_k', z\big),
\end{align*}
and $\sim_e$ is a temporally abstract bisimulation.
\end{lemma}

\begin{theorem}[Global optima of $\mathcal{L}$ yield temporally abstract bisimulation]
\label{thm:elbo-tab}
Let $P_\Phi(x, \tilde{s}, z) = p_{\mathrm{data}}(x)\, q_\Phi(\tilde{s}, z \mid x)$ denote the true joint distribution of
trajectories and inferred latents. Assume the model classes are sufficiently expressive that
$p_\Theta(x, \tilde{s}, z) = P_\Phi(x, \tilde{s}, z)$ for some parameters $(\Theta, \Phi)$. Then any global maximizer $(\Theta, \Phi)$ of $\mathcal{L}(\Theta, \Phi)$ satisfies, for all
$k$ and $(h_k, \tilde{s}_k, z_k)$ with positive probability under $P_\Phi$,
\[
    p_\psi(\tilde{s}_{k+1} \mid \tilde{s}_k, z_k) = P_\Phi(\tilde{s}_{k+1} \mid h_k, \tilde{s}_k, z_k),
    \qquad
    p_\rho(\tilde{r}_k \mid \tilde{s}_k, z_k) = P_\Phi(\tilde{r}_k \mid h_k, \tilde{s}_k, z_k).
\]
These equalities imply the dynamics \eqref{eq:dyn-suff} and reward \eqref{eq:rew-suff} sufficiency conditions, and therefore, by
Lemma~\ref{lem:suff-tab}, the encoder-induced partition of histories is a temporally abstract
bisimulation.
\end{theorem}

Maximizing $\mathcal{L}(\Theta, \Phi) = -\mathcal{H}(p_{\mathrm{data}}) - \mathrm{KL}\big[P_\Phi \,\|\, p_\Theta\big]$ is equivalent to minimizing $\mathrm{KL}\big[P_\Phi \,\|\, p_\Theta\big]$, the divergence between the true distribution of trajectories and inferred latents and the one generated by the model. Under realizability the optimum makes the two equal. In the model, the reward and dynamics heads see only $(\tilde{s}_k, z_k)$, so the history cannot affect the segment reward or the next abstract state beyond what $(\tilde{s}_k, z_k)$ already carries. Since the true distribution equals the model's, the same holds in the data, which is reward and dynamics sufficiency. A complete proof is in Appendix~\ref{app:proof-elbo-tab}.

\subsection{Planning with TSAWM}
Once the world model and goal-conditioned reward function is learned, the agent plans over temporally extended skills in the learned abstract state space. Given a current abstract state \(\tilde{s}_0\) and goal \(g\), we use the cross-entropy method (CEM) \citep{Rubinstein1999TheCM} to optimize a sequence of normalized skill variables \(\epsilon_{0:L-1}\). Each \(\epsilon_\ell\) is mapped to a skill through the state-conditioned prior, $z_k = \mu_\omega(\tilde{s}_\ell) + \sigma_\omega(\tilde{s}_\ell)\odot \epsilon_k$, where $p_\omega(z_\ell\mid \tilde{s}_\ell)=\mathcal{N}(\mu_\omega(\tilde{s}_\ell),\sigma_\omega(\tilde{s}_\ell))$.
Planning in this whitened space gives a better-conditioned search problem, since likely skills correspond to values near a unit Gaussian. Candidate skill sequences are then rolled out through the abstract dynamics and scored using the trained goal-conditioned reward decoder. We use model predictive control where, after CEM selects a sequence, the agent executes only the first skill through the low-level policy for \(H\) steps before replanning. This makes planning more robust to errors in the abstract dynamics model. The hyperparameters used for CEM planning are detailed in the Appendix in Table \ref{tab:hparams}.

\section{Experiments}
Our central hypothesis is that there exists a synergistic relationship between temporal and state abstraction, where planning over longer skills obviates the need for the state to keep information needed for fine-grained control, effectively reducing the state information required by the planner. To that end, we test this in three ways. First, we vary the information encoded in the abstract states using an information bottleneck along with the skill length $H$, and measure the planning success in varying settings. Second, we probe the components of the state that are discarded by the encoder as $H$ increases. Finally, we compare our method against both model-free and model-based baselines in OGBench. 

\subsection{Experimental Domains}
\label{sec:exp-domains}
In this work, we explore environments from both D4RL \citep{fu2020d4rl} and OGBench \citep{park2025ogbench}. From D4RL, we focus on the \textit{medium} layouts of the \textit{Maze2D} and \textit{AntMaze} environments. The D4RL environments are used only for the experiments to understand the relationship between temporal and state abstraction as its locomotion mazes are simpler than the OGBench tasks. This makes them a natural sandbox for studying the synergy between the two forms of abstraction. We use OGBench to compare our model with existing offline goal-conditioned methods. We select the three state-based locomotion domains, \textit{pointmaze} (\textit{medium}, \textit{large}), \textit{antmaze} (\textit{medium}, \textit{large}), and \textit{humanoidmaze} (\textit{medium}). For manipulation, we use \textit{cube-single}. In every domain, we use the goal-conditioned datasets (\texttt{navigate} for the mazes, \texttt{play} for cube) rather than the single-task variants. We defer further information about each environment to the appendix.

\subsection{Skill Horizon and the State Information Required for Planning}
\label{sec:exp-rate}

A coarser state abstraction groups more histories together, so it can be represented with fewer bits. We test the empirical result of this claim: \textit{when longer skills can be built by composing shorter ones, an agent with longer skills should maintain planning success even with less information encoded in the abstract state}. We choose skill lengths of $H \in \{5, 10, 20\}$ such that all lengths share a common divisor of 5, which makes it possible for the longer skills to be composed of shorter ones. However, each model is trained separately; thus, longer skills are not forced to be compositions. Whether the predicted ordering in abstraction coarseness appears is the empirical question we would like to answer. 

\paragraph{Limiting the information in the abstract state.}
We limit the information in $\tilde{s}_k$ directly through the encoder using an information bottleneck. Each of the $d_{\tilde{s}}$ dimensions of the posterior mean is squashed into $[-M, M]$ using $M\tanh(\cdot/M)$ and the posterior standard deviation is fixed to a constant $\sigma$. Because the mean is bounded and the noise is fixed, $\tilde{s}_k$ can carry at most $\frac{d_{\tilde{s}}}{2}\log_2(1 + M^2/\sigma^2)$ bits. We use $M = 1$ and sweep $\sigma \in \{0.5, 0.3, 0.2, 0.1, 0.01, 0.001\}$, where lowering $\sigma$ raises the upper bound on information.

We vary the skill length $H \in \{5, 10, 16, 20\}$ on D4RL Maze2D-Medium and AntMaze-Medium-Diverse \citep{fu2020d4rl}. To isolate the skill horizon, every model is trained on the same $80$-step windows of
low-level observations and for the same $10^6$ gradient
steps. All other hyperparameters are shared. At test time the planner optimizes $L=K$ skills with the same $80$-step lookahead for every $H$, using the segment reward head. We train $3$ seeds per $(\sigma, H)$ combination and evaluate each on $100$ episodes.

\begin{figure}[t]
  \centering
  \begin{subfigure}[t]{0.49\linewidth}
    \centering
    \includegraphics[width=\linewidth]{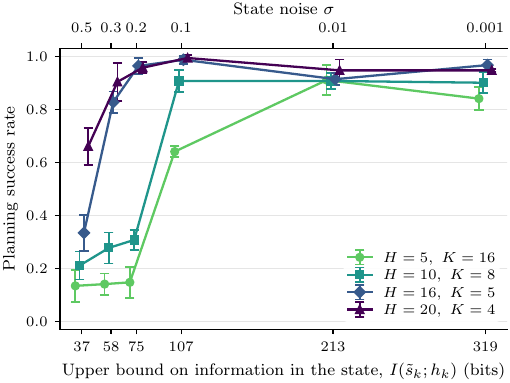}
    \caption{Maze2D-Medium}
    \label{fig:rate-maze2d}
  \end{subfigure}\hfill
  \begin{subfigure}[t]{0.49\linewidth}
    \centering
    \includegraphics[width=\linewidth]{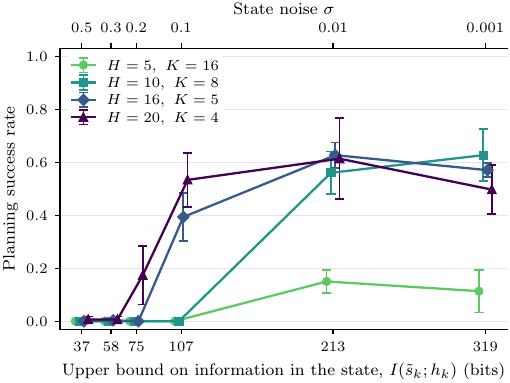}
    \caption{AntMaze-Medium-Diverse}
    \label{fig:rate-antmaze}
  \end{subfigure}
  \caption{Planning success against the information allowed in the abstract state, for four skill lengths. The bottom axis is the upper bound on $I(\tilde{s}_k; h_k)$ and the top axis is the corresponding noise $\sigma$. Points are the mean over $3$ seeds ($100$ episodes each) and error bars are one standard deviation across seeds. Longer skills keep their success even for larger standard deviations.}
  \label{fig:rate-horizon}
\end{figure}

Figure~\ref{fig:rate-horizon} shows the result. With a higher upper bound on the information in the abstract state, planning success is roughly the same for every $H$. As the upper bound tightens, success falls in order of skill length from shortest ($H=5$) to longest ($H=20$). AntMaze shows a similar trend, but shifted right towards more information. These results are consistent with our theoretical prediction. Along the chain $5 \Rightarrow 10 \Rightarrow 20$, the model is capable of learning skills that are compositions of the shorter skills. Additionally, the abstract state requires less information while preserving planning success. This indicates that the abstraction coarsens when a longer skill can be a composition of shorter ones.

\subsection{What Information the Abstract State Discards as Skills Lengthen}
\label{sec:exp-content}

Section~\ref{sec:exp-rate} showed that longer skills maintain planning success with less information in $\tilde{s}_k$. We now probe exactly what information is given up by looking inside the learned abstract states of the AntMaze models and measuring which parts of the ant's state are kept at each skill length. From the $29$-dimensional ant observation, we form seven target groups: position $(x, y)$; heading (the cosine and sine of the torso's yaw); torso height and uprightness (the vertical component of the torso's up axis); joint angles (8 in total); the torso's $xy$ velocity; the torso's angular velocity; and joint velocities.

We fit an MLP probe from $\tilde{s}_k$ to each group. The probe has two hidden layers of dimension $256$ with ReLU activations and is trained with Adam (learning rate $10^{-3}$, weight decay $10^{-4}$, batch size $2048$, at most $3000$ steps). We report the held-out $R^2$. We also estimate how much information each abstract state actually carries. For each dimension of the abstract state, $I(\tilde{s}_{k,d}; h_k)$ is the entropy of $\tilde{s}_{k,d}$ minus the entropy of the noise $\sigma \epsilon_d$, where the first entropy is estimated with $8$ noise draws per state. The sum over dimensions is an upper bound on $I(\tilde{s}_k; h_k)$.

\begin{figure}[t]
  \centering
  \includegraphics[width=\linewidth]{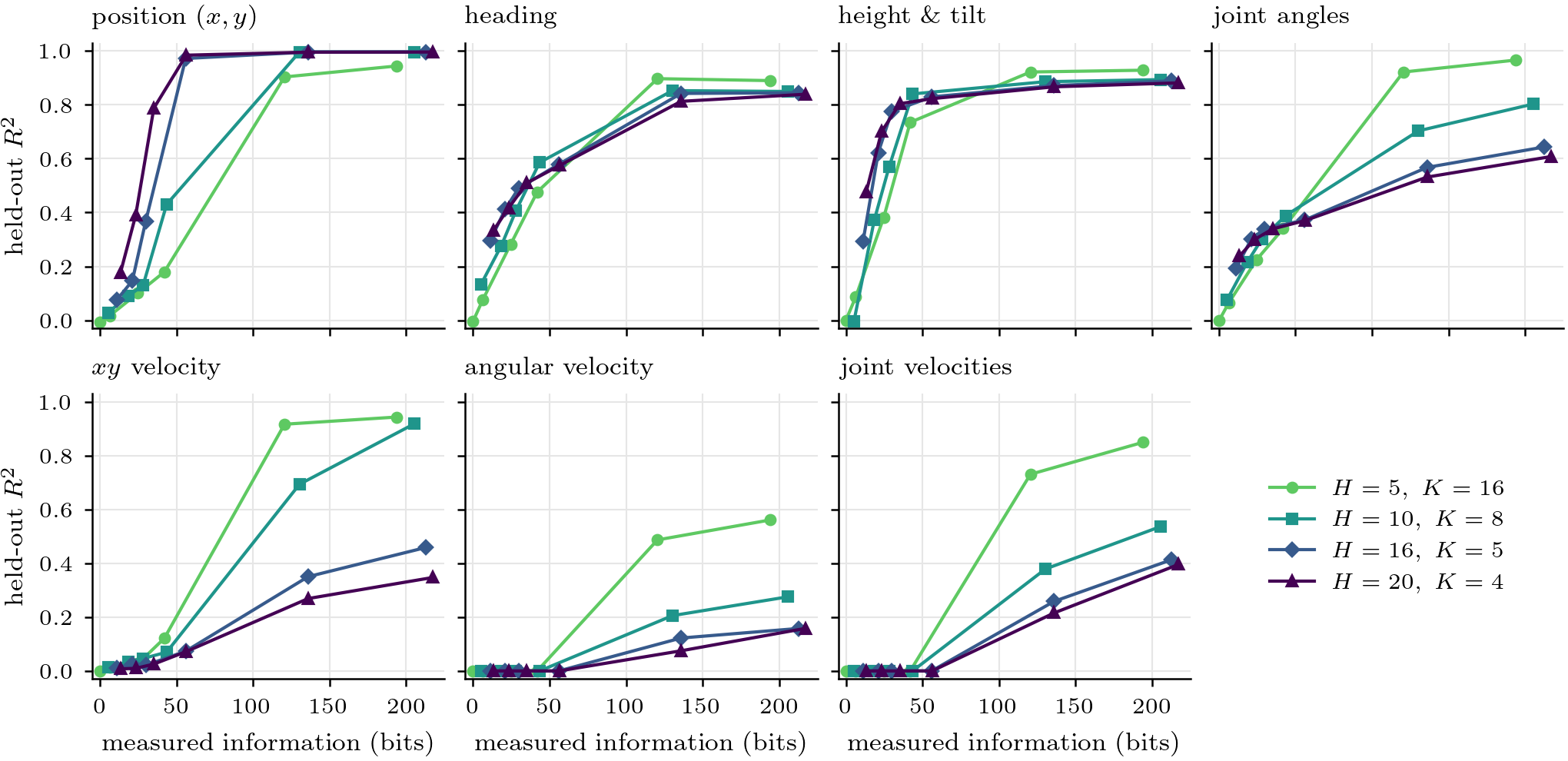}
  \caption{Held-out $R^2$ of each group against the measured information in the abstract state for each skill length. Each point is one value of $\sigma$, with both coordinates averaged over $3$ seeds. At the same number of
  bits, longer skills put more of the abstract state into position and less into
  the joint angles and velocities.}
  \label{fig:content-bits}
\end{figure}

\paragraph{Longer skills keep position and drop the body state.}
In Figure~\ref{fig:content-bits}, at the same number of bits, longer skills reach a higher $R^2$ for position and a lower one for groups like angular velocity and joint velocities. For example, $H = 20$ with a stricter information bottleneck of $\sigma = 0.1$ carries only $56$ bits, fewer than half of the $121$ bits that $H = 5$ encodes with a looser bottleneck of $\sigma = 0.01$. Yet the $H = 20$ states decode position more accurately ($R^2 = 0.98$ versus $R^2 = 0.90$). This confirms our intuition that a longer skill reduces the need for information relevant for fine-grained control to be encoded in the state. As skills lengthen, the learned abstraction stops distinguishing histories that differ only in body state, whose effect fades within a skill. It keeps the slowly varying quantities that are paramount to the segment outcome instead, including position, heading and uprightness.

\begin{table}[t]
  \centering
  \small
  \begin{tabular}{lccccccc}
    \toprule
    Environment (dataset) & GCBC & GCIVL & GCIQL & QRL & CRL & HIQL & Ours \\
    \midrule
    \texttt{antmaze-medium-navigate}      & $29 \pm 4$ & $72 \pm 8$ & $71 \pm 4$ & $88 \pm 3$ & $95 \pm 1$ & $\mathbf{96 \pm 1}$ & $77 \pm 5$ \\
    \texttt{antmaze-large-navigate}       & $24 \pm 2$ & $16 \pm 5$ & $34 \pm 4$ & $75 \pm 6$ & $83 \pm 4$ & $\mathbf{91 \pm 2}$ & $49 \pm 5$ \\
    \texttt{humanoidmaze-medium-navigate} & $8 \pm 2$  & $24 \pm 2$ & $27 \pm 2$ & $21 \pm 8$ & $60 \pm 4$ & $\mathbf{89 \pm 2}$ & $45 \pm 4$ \\
    \texttt{pointmaze-medium-navigate}    & $9 \pm 6$  & $63 \pm 6$ & $53 \pm 8$ & $82 \pm 5$ & $29 \pm 7$ & $79 \pm 5$ & $\mathbf{95 \pm 3}$ \\
    \texttt{pointmaze-large-navigate}     & $29 \pm 6$ & $45 \pm 5$ & $34 \pm 3$ & $\mathbf{86 \pm 9}$ & $39 \pm 7$ & $58 \pm 5$ & $69 \pm 7$ \\
    \texttt{cube-single-play}             & $6 \pm 2$  & $53 \pm 4$ & $\mathbf{68 \pm 6}$ & $5 \pm 1$ & $19 \pm 2$ & $15 \pm 3$ & $25 \pm 4$ \\
    \midrule
    Average                               & $17.5$ & $45.5$ & $47.8$ & $59.5$ & $54.2$ & $\mathbf{71.3}$ & $60.0$ \\
    Average rank $\downarrow$             & $6.7$  & $4.5$  & $4.5$  & $3.7$  & $3.5$  & $\mathbf{2.3}$  & $2.8$ \\
    \bottomrule
  \end{tabular}
  \caption{Success rate (\%) on OGBench averaged over $8$ seeds with the standard deviation after the $\pm$ sign. Baseline numbers are from \citet{park2025ogbench}. Our method uses no
  value function, no shaped reward and no graph. Following OGBench, each seed's success is the mean over its checkpoints at $800$K, $900$K and $1$M gradient steps, each evaluated on $250$ episodes ($50$ per task). Bold denotes values the largest mean in each row. Average is the mean success over the six environments; average rank is the mean of each method's rank in each environment ($1$ = best, ties share the mean
  rank).}
  \label{tab:ogbench-baselines}
\end{table}

\section{Comparison with Model-Free and Model-Based Baselines}
We compare against the six model-free offline goal-conditioned RL methods evaluated by OGBench \citep{park2025ogbench}. All baselines are trained on the same datasets with hindsight-relabeled goals for $10^6$ gradient steps. GCBC \citep{Ghosh2019LearningTR} is goal-conditioned behavioral cloning that imitates the dataset action using a future state of the same trajectory as the goal. GCIVL \citep{gcivl} and GCIQL \citep{gciql} fit the optimal goal-conditioned value by expectile regression on the sparse goal reward, in the style of IQL \citep{gcivl}. QRL \citep{qrl} learns the value as a quasimetric distance function that obeys the triangle inequality. CRL \citep{crl} uses contrastive learning to learn a goal-conditioned value function. HIQL \citep{hiql} learns a single GCIVL-style value and extracts a two-level policy from it: a high-level policy proposes a subgoal and a low-level policy reaches it.

For the model-based comparison, we use ALPS \citep{alps2026laplacian}, a recent model-based method evaluated in the OGBench environments. ALPS learns a Laplacian representation of the original state space that captures the commute time between different states. It clusters states in this representation, and chains together cluster centers as subgoals with Dijkstra's algorithm. Then, it uses CEM to reach each subgoal using a one-step model in the original state space seeded by a goal-conditioned behavior prior.

As seen in Table \ref{tab:ogbench-baselines}, our method is competitive with the model-free baselines. It achieves the second-best average rank ($2.8$, behind HIQL's $2.3$) and second-highest average planning success ($60\%$, behind HIQL's $71.3\%$) across the evaluated environments. It also scores the highest on the \texttt{pointmaze-medium-navigate} environment. This is significant because our approach does not learn a value function and plans only with a learned prediction of the sparse environment reward. Although ALPS achieves the highest planning success in $4$ out of $5$ environments, the boost in long-horizon performance comes from graph search over clusters. When comparing ALPS's planner without a graph, CEM with a behavior prior, which is the closest analogue to ours, our approach wins in planning success in all $5$ environments.

\section{Discussion}
This paper presents a new representation learning method for jointly learning temporal and state abstractions. Learning a skill set alongside the abstract states directly shapes the coarseness of the state abstraction, which we formalize through a \emph{temporally abstract bisimulation}. We show that in maze environments, learning longer skills encourages the abstract state to discard body state information while preserving planning-relevant details like position and uprightness. This allows the high-level planner to plan over states with less information. Finally, our method achieves similar performance on long-horizon tasks without requiring learning a value function for planning, unlike other methods.

\begin{table}[t]
  \centering
  \small
  \begin{tabular}{lcccc}
    \toprule
                                          & \multicolumn{2}{c}{Without graph} & With graph \\
    \cmidrule(lr){2-4} \cmidrule(lr){5-5}
    Environment (dataset)                 & ALPS CEM + prior & Ours & ALPS (full) \\
    \midrule
    \texttt{antmaze-medium-navigate}      & $57 \pm 8$  & $\mathbf{77 \pm 5}$ & $97 \pm 2$  \\
    \texttt{antmaze-large-navigate}       & $29 \pm 12$ & $\mathbf{49 \pm 5}$ & $93 \pm 5$  \\
    \texttt{humanoidmaze-medium-navigate} & $26 \pm 10$ & $\mathbf{45 \pm 4}$ & $89 \pm 5$  \\
    \texttt{pointmaze-medium-navigate}    & $30 \pm 7$  & $\mathbf{95 \pm 3}$ & $82 \pm 10$ \\
    \texttt{pointmaze-large-navigate}     & $21 \pm 9$  & $\mathbf{69 \pm 7}$ & $80 \pm 8$  \\
    \bottomrule
  \end{tabular}
  \caption{Planning without a graph (numbers from \citet{alps2026laplacian})  evaluated with the same protocol as Table~\ref{tab:ogbench-baselines}. \emph{CEM + prior} is its flat planner, CEM with a one-step model in raw state space intialized by the
  prior. Full ALPS adds Dijkstra search over a graph of clusters in its Laplacian representation. Ours plans with CEM over $15$ skills in the learned abstract model, with no graph and no goal-conditioned prior. Bold marks the best
  method without a graph.}
  \label{tab:flat-planning}
\end{table}

\newpage
\bibliography{main}
\bibliographystyle{tmlr}
\newpage
\appendix

\section{Derivation of the Model Objective}
\label{app:derivation}

\paragraph{Notation.}
We write $x = (o_{0:T}, a_{0:T-1}, \tilde{r}_{0:K-1})$ for a trajectory, $\tilde{s} = \tilde{s}_{0:K}$ and
$z = z_{0:K-1}$ for the latents, $p_{\mathrm{data}}$ for the true distribution of trajectories, and
$p_\Theta$ for the generative model \eqref{eq:generative}.

The marginal likelihood of a trajectory integrates the latents out of the generative model. Multiplying and dividing by the posterior $q_\Phi(\tilde{s}, z \mid x)$ and applying Jensen's inequality,
\begin{align}
    \log p_\Theta(x)
      &= \log \int p_\Theta(x, \tilde{s}, z)\, d\tilde{s}\, dz
       = \log \int q_\Phi(\tilde{s}, z \mid x)\, \frac{p_\Theta(x, \tilde{s}, z)}{q_\Phi(\tilde{s}, z \mid x)}\, d\tilde{s}\, dz \notag\\
      &\ge \mathbb{E}_{q_\Phi(\tilde{s}, z \mid x)}\Big[\log \frac{p_\Theta(x, \tilde{s}, z)}{q_\Phi(\tilde{s}, z \mid x)}\Big].
    \label{eq:jensen}
\end{align}
We now factor the numerator and the denominator.

\paragraph{Numerator.}
Taking the logarithm of the generative model \eqref{eq:generative},
\begin{align}
    \log p_\Theta(x, \tilde{s}, z)
      &= \log p(\tilde{s}_0) + \log p_{\eta_0}(o_0 \mid \tilde{s}_0)
       + \sum_{k=0}^{K-1} \Big[ \log p_\omega(z_k \mid \tilde{s}_k)
       + \log p_\rho(\tilde{r}_k \mid \tilde{s}_k, z_k) \notag\\
      &\qquad
       + \log p_\psi(\tilde{s}_{k+1} \mid \tilde{s}_k, z_k)
       + \log p_\eta(\bar{o}_k \mid \tilde{s}_k, z_k, \tilde{s}_{k+1}, h_k)
       + \sum_{t=t_k}^{t_{k+1}-1} \log \pi_\theta(a_t \mid o_t, z_k) \Big].
    \label{eq:log-num}
\end{align}

\paragraph{Denominator.}
Taking the logarithm of the posterior \eqref{eq:posterior},
\begin{equation}
    \log q_\Phi(\tilde{s}, z \mid x)
      = \log q_\chi(\tilde{s}_0 \mid h_0)
      + \sum_{k=0}^{K-1} \Big[ \log q_\phi(z_k \mid \bar{o}_k, \bar{a}_k, \tilde{s}_k)
      + \log q_\chi(\tilde{s}_{k+1} \mid h_{k+1}) \Big].
    \label{eq:log-den}
\end{equation}

Subtracting \eqref{eq:log-den} from \eqref{eq:log-num} and grouping the terms that involve the same latent
variable,
\begin{align}
    \log \frac{p_\Theta(x, \tilde{s}, z)}{q_\Phi(\tilde{s}, z \mid x)}
      &= \log \frac{p(\tilde{s}_0)}{q_\chi(\tilde{s}_0 \mid h_0)} + \log p_{\eta_0}(o_0 \mid \tilde{s}_0)
       + \sum_{k=0}^{K-1} \Big[
         \log \frac{p_\omega(z_k \mid \tilde{s}_k)}{q_\phi(z_k \mid \bar{o}_k, \bar{a}_k, \tilde{s}_k)}
       + \log \frac{p_\psi(\tilde{s}_{k+1} \mid \tilde{s}_k, z_k)}{q_\chi(\tilde{s}_{k+1} \mid h_{k+1})} \notag\\
      &\qquad
       + \log p_\rho(\tilde{r}_k \mid \tilde{s}_k, z_k)
       + \log p_\eta(\bar{o}_k \mid \tilde{s}_k, z_k, \tilde{s}_{k+1}, h_k)
       + \sum_{t=t_k}^{t_{k+1}-1} \log \pi_\theta(a_t \mid o_t, z_k) \Big].
    \label{eq:log-ratio}
\end{align}

The observed variables $o_0$, $\tilde{r}_k$, $\bar{o}_k$ and $a_t$ appear only in the numerator, so they remain as log-likelihood terms. Each latent variable is grouped into a prior--posterior pair, so taking the expectation over $q_\Phi$ changes each pair into a KL term. The paired terms depends on only a few latents, so its expectation only needs their marginal under $q_\Phi$.
Integrating the remaining latents out of \eqref{eq:posterior}, the factors of later latents integrate to
one, the factors of earlier latents integrate to one, and $\tilde{s}_k$ depends on $x$ only through $h_k$, so
\begin{equation}
    q_\Phi(\tilde{s}_k, z_k, \tilde{s}_{k+1} \mid x)
      = q_\chi(\tilde{s}_k \mid h_k)\; q_\phi(z_k \mid \bar{o}_k, \bar{a}_k, \tilde{s}_k)\; q_\chi(\tilde{s}_{k+1} \mid h_{k+1}).
    \label{eq:q-marg}
\end{equation}
For the next abstract state, \eqref{eq:q-marg} gives
\begin{align}
    \mathbb{E}_{q_\Phi}\Big[\log \frac{p_\psi(\tilde{s}_{k+1} \mid \tilde{s}_k, z_k)}{q_\chi(\tilde{s}_{k+1} \mid h_{k+1})}\Big]
      &= \int q_\chi(\tilde{s}_k \mid h_k)\, q_\phi(z_k \mid \bar{o}_k, \bar{a}_k, \tilde{s}_k) \notag\\
      &\qquad \times \Big[ \int q_\chi(\tilde{s}_{k+1} \mid h_{k+1})
         \log \frac{p_\psi(\tilde{s}_{k+1} \mid \tilde{s}_k, z_k)}{q_\chi(\tilde{s}_{k+1} \mid h_{k+1})}\, d\tilde{s}_{k+1} \Big]
         d\tilde{s}_k\, dz_k \notag\\
      &= -\,\mathbb{E}_{q_\Phi}\Big[ \mathrm{KL}\big[\, q_\chi(\tilde{s}_{k+1} \mid h_{k+1}) \,\big\|\,
         p_\psi(\tilde{s}_{k+1} \mid \tilde{s}_k, z_k) \,\big] \Big].
    \label{eq:kl-S}
\end{align}
The inner integral is a negative KL divergence because $\tilde{s}_{k+1}$ is drawn from $q_\chi(\cdot \mid h_{k+1})$
independent of $(\tilde{s}_k, z_k)$, so the posterior factor can be pulled inside. For the skill, the paired term
depends only on $(\tilde{s}_k, z_k)$, and
\begin{align}
    \mathbb{E}_{q_\Phi}\Big[\log \frac{p_\omega(z_k \mid \tilde{s}_k)}{q_\phi(z_k \mid \bar{o}_k, \bar{a}_k, \tilde{s}_k)}\Big]
      &= \int q_\chi(\tilde{s}_k \mid h_k)
         \Big[ \int q_\phi(z_k \mid \bar{o}_k, \bar{a}_k, \tilde{s}_k)
         \log \frac{p_\omega(z_k \mid \tilde{s}_k)}{q_\phi(z_k \mid \bar{o}_k, \bar{a}_k, \tilde{s}_k)}\, dz_k \Big] d\tilde{s}_k \notag\\
      &= -\,\mathbb{E}_{q_\Phi}\Big[ \mathrm{KL}\big[\, q_\phi(z_k \mid \bar{o}_k, \bar{a}_k, \tilde{s}_k) \,\big\|\,
         p_\omega(\cdot \mid \tilde{s}_k) \,\big] \Big].
    \label{eq:kl-z}
\end{align}
For the initial state, the paired term depends only on $\tilde{s}_0$, whose posterior depends only on $x$, so
\begin{equation}
    \mathbb{E}_{q_\Phi}\Big[\log \frac{p(\tilde{s}_0)}{q_\chi(\tilde{s}_0 \mid h_0)}\Big]
      = -\,\mathrm{KL}\big[\, q_\chi(\tilde{s}_0 \mid h_0) \,\big\|\, p(\tilde{s}_0) \,\big].
    \label{eq:kl-S0}
\end{equation}

Taking the expectation of \eqref{eq:log-ratio} under $q_\Phi$, and replacing the three paired terms by \eqref{eq:kl-S}, \eqref{eq:kl-z} and \eqref{eq:kl-S0}, the right-hand side of \eqref{eq:jensen} becomes
exactly $\mathcal{L}(\Theta, \Phi; x)$ in \eqref{eq:elbo}. Averaging over the data gives $\mathcal{L}(\Theta, \Phi) = \mathbb{E}_{x \sim p_{\mathrm{data}}}[\mathcal{L}(\Theta, \Phi; x)]$.

\section{Proofs from Section \ref{sec:TAB}}

\subsection{Proof of Theorem~\ref{thm:optimal-control}}
\label{app:optimal}
\begin{proof}
We assume rewards are bounded and $\gamma \in [0,1)$, and every history has at least one available option. We use a bar to denote quantities of the quotient $\mathcal{M}/\!\sim$. For a class $C$ of $\sim$ and an option $z\in \mathcal{Z}_C$, let
\[
    \bar{R}(C, z) = \mathbb{E}[\tilde{r}_k \mid h_k, z],
    \qquad
    \bar{K}_\gamma(C' \mid C, z) = K_\gamma(C' \mid h_k, z)
\]
for any $h_k \in C$. By \eqref{eq:avail}, $z$ is available at every $h_k \in C$, and by \eqref{eq:rew-suff-gen} and \eqref{eq:dyn-suff-gen}, these quantities do not depend on which $h_k \in C$ is chosen. The quotient $\mathcal{M}/\!\sim$ is the semi-MDP whose states are the classes of $\sim$, whose options at class $C$ are $\mathcal{Z}_C$, and whose reward and discounted transitions are $\bar{R}$ and $\bar{K}_\gamma$. The optimal values of $\mathcal{M}$ and $\mathcal{M}/\!\sim$ satisfy the semi-MDP Bellman equations \citep{sutton1999between},
\begin{align*}
    V^*(h_k) &= \sup_{z \in \mathcal{Z}(h_k)} \Big( \mathbb{E}[\tilde{r}_k \mid h_k, z] + \mathbb{E}\big[\gamma^{\tau_k}\, V^*(h_{k+1}) \mid h_k, z\big] \Big), \\
    \bar{V}^*(C) &= \sup_{z \in \mathcal{Z}_C} \Big( \bar{R}(C, z) + \sum_{C'} \bar{K}_\gamma(C' \mid C, z)\, \bar{V}^*(C') \Big).
\end{align*}

We first show that a history has the same optimal value as its class. Let $V_0 = 0$ and $\bar{V}_0 = 0$, and obtain $V_{n+1}$ and $\bar{V}_{n+1}$ by applying the right-hand sides above to $V_n$ and $\bar{V}_n$. We show by induction on $n$ that $V_n(h_k) = \bar{V}_n(C)$ for every class $C$ and every $h_k \in C$. The base case holds since both are zero. Suppose it holds for $n$. Then for every $h_k \in C$ and every option $z \in \mathcal{Z}_C$,
\begin{align*}
    \mathbb{E}\big[\gamma^{\tau_k}\, V_n(h_{k+1}) \mid h_k, z\big]
      &= \sum_{C'} \mathbb{E}\big[\gamma^{\tau_k}\, \mathbf{1}\{h_{k+1} \in C'\}\, V_n(h_{k+1}) \mid h_k, z\big] \\
      &= \sum_{C'} \mathbb{E}\big[\gamma^{\tau_k}\, \mathbf{1}\{h_{k+1} \in C'\} \mid h_k, z\big]\, \bar{V}_n(C')
         && \text{(By the induction hypothesis)} \\
      &= \sum_{C'} \bar{K}_\gamma(C' \mid C, z)\, \bar{V}_n(C')
         && \text{(By \eqref{eq:dyn-suff-gen})}.
\end{align*}
Together with $\mathbb{E}[\tilde{r}_k \mid h_k, z] = \bar{R}(C, z)$ from \eqref{eq:rew-suff-gen}, the terms inside the two suprema are equal for every option in $\mathcal{Z}_C$. By \eqref{eq:avail}, $\mathcal{Z}(h_k) = \mathcal{Z}_C$, so the two suprema have the same initiation set and $V_{n+1}(h_k) = \bar{V}_{n+1}(C)$, which completes the induction. Let $T$ and $\bar{T}$ denote the updates $V_n \mapsto V_{n+1}$ and $\bar{V}_n \mapsto \bar{V}_{n+1}$. Since every option takes at least one step, $\tau_k \ge 1$, so $\gamma^{\tau_k} \le \gamma$ and hence $\mathbb{E}[\gamma^{\tau_k} \mid h_k, z] \le \gamma$. For bounded functions $V$ and $W$ on histories and any option $z \in \mathcal{Z}(h_k)$, the reward terms cancel, so
\[
    \Big| \mathbb{E}\big[\gamma^{\tau_k} V(h_{k+1}) \mid h_k, z\big] - \mathbb{E}\big[\gamma^{\tau_k} W(h_{k+1}) \mid h_k, z\big] \Big|
    \le \mathbb{E}\big[\gamma^{\tau_k} \mid h_k, z\big]\, \|V - W\|_\infty
    \le \gamma\, \|V - W\|_\infty .
\]
Taking the supremum over options gives $\|TV - TW\|_\infty \le \gamma \|V - W\|_\infty$. The same argument applies to $\bar{T}$, because $\sum_{C'} \bar{K}_\gamma(C' \mid C, z) = \mathbb{E}[\gamma^{\tau_k} \mid h_k, z] \le \gamma$ for any $h_k \in C$. Since $V^*$ and $\bar{V}^*$ are fixed points of $T$ and $\bar{T}$, $\|V_n - V^*\|_\infty \le \gamma^n \|V_0 - V^*\|_\infty \to 0$, and likewise $\bar{V}_n \to \bar{V}^*$. Taking $n \to \infty$ in $V_n(h_k) = \bar{V}_n(C)$ gives $V^*(h_k) = \bar{V}^*(C)$ for every $h_k \in C$. Hence $V_n \to V^*$ and $\bar{V}_n \to \bar{V}^*$, and $V^*(h_k) = \bar{V}^*(C)$ for every $h_k \in C$.

Next, let $Q^*(h_k, z) = \mathbb{E}[\tilde{r}_k \mid h_k, z] + \mathbb{E}[\gamma^{\tau_k}\, V^*(h_{k+1}) \mid h_k, z]$ and $\bar{Q}^*(C, z) = \bar{R}(C, z) + \sum_{C'} \bar{K}_\gamma(C' \mid C, z)\, \bar{V}^*(C')$. Repeating the computation above with $V^*$ in place of $V_n$ gives $Q^*(h_k, z) = \bar{Q}^*(C, z)$ for every $h_k \in C$ and every option $z$.

For each history $h_k$, let $C$ be its class and define $\pi(\cdot \mid h_k) = \bar{\pi}(\cdot \mid C)$, which only selects options available at $h_k$, since $\mathcal{Z}_C = \mathcal{Z}(h_k)$. Then
\[
    \mathbb{E}_{z \sim \pi(\cdot \mid h_k)}\big[Q^*(h_k, z)\big]
    = \mathbb{E}_{z \sim \bar{\pi}(\cdot \mid C)}\big[\bar{Q}^*(C, z)\big]
    = \bar{V}^*(C) = V^*(h_k),
\]
so $V^*$ satisfies the Bellman equation of $\pi$. That equation is a $\gamma$-contraction with the unique fixed point $V^\pi$, so $V^\pi = V^*$ and $\pi$ is optimal on $\mathcal{M}$.
\end{proof}

\subsection{Proof of Theorem~\ref{thm:coarsest}}
\label{app:coarsest}

\begin{proof}
By Definition~\ref{def:bisim}, an equivalence relation $\sim$ is a TAB exactly when, for all $h_k \sim h_k'$, $\mathcal{Z}(h_k) = \mathcal{Z}(h_k')$ and, for every option $z_k \in \mathcal{Z}(h_k)$,
\[
    K_\gamma(C \mid h_k, z_k) = K_\gamma(C \mid h_k', z_k) \;\; \text{for every class } C \text{ of } \sim,
    \qquad
    \mathbb{E}[\tilde{r}_k \mid h_k, z_k] = \mathbb{E}[\tilde{r}_k \mid h_k', z_k].
\]

We now construct the coarsest TAB, $\sim_{\mathrm{TA}}$. Let $h \sim_{\mathrm{TA}} h'$ if there is a finite chain $h = h^{(0)}, h^{(1)}, \ldots, h^{(n)} = h'$ where each consecutive history is related by some TAB. It is also clear that $\sim_{\mathrm{TA}}$ is an equivalence relation, because every TAB relates a history to itself and chains can be reversed and joined. Additionally, this equivalence relation contains every TAB, since each pair of histories considered equivalent by a TAB is a chain with $n = 1$.

We now show that $\sim_{\mathrm{TA}}$ is a TAB. First take a temporally abstract bisimulation $\sim$, histories $h_k \sim h_k'$, and an option $z_k \in \mathcal{Z}(h_k)$. Because $\sim_{\mathrm{TA}}$ contains $\sim$, every class $C$ of $\sim_{\mathrm{TA}}$ is a union of classes $C_1, C_2, \ldots$ of $\sim$. Since $\mathbf{1}\{h_{k+1} \in C\} = \sum_i \mathbf{1}\{h_{k+1} \in C_i\}$,
\[
    K_\gamma(C \mid h_k, z_k) = \sum_{i} K_\gamma(C_i \mid h_k, z_k) = \sum_{i} K_\gamma(C_i \mid h_k', z_k) = K_\gamma(C \mid h_k', z_k),
\]
where the middle equality holds because $\sim$ is a TAB. The availability and reward equalities hold directly because they do not depend on the classes. Now take any $h_k \sim_{\mathrm{TA}} h_k'$. Each consecutive pair in the chain from $h_k$ to $h_k'$ is related by some TAB, so the availability equality holds across each pair, and every option available at $h_k$ is available at every history in the chain. The dynamics and reward equalities, with the classes of $\sim_{\mathrm{TA}}$, therefore hold across each pair for these options. By transitivity of equality, all three hold for $h_k$ and $h_k'$. So $\sim_{\mathrm{TA}}$ is a TAB.

Since $\sim_{\mathrm{TA}}$ is a TAB that contains every TAB, it is the coarsest. It is unique because two coarsest TABs would contain each other and therefore be equal.
\end{proof}

\subsection{Proof of Theorem~\ref{thm:composition}}
\label{app:composition}

\begin{proof}
For a temporally abstract bisimulation, we write $\bar{R}(C, z)$ and $\bar{K}_\gamma(C' \mid C, z)$ for the common values of $\mathbb{E}[\tilde{r}_k \mid h_k, z]$ and $K_\gamma(C' \mid h_k, z)$ over $h_k \in C$, as in Appendix~\ref{app:optimal}. We prove by induction on $m$ that $\sim$ is a temporally abstract bisimulation for $\mathcal{Z}^{(m)}$. For $m = 1$, this is the hypothesis.

For the inductive step, write a composition of $m+1$ skills as $z_1 z_2$, where $z_1 \in \mathcal{Z}^{(m)}$ is the composition of the first $m$ skills and $z_2 \in \mathcal{Z}$ is the last one. We assume throughout that the events conditioned on have positive probability. We index boundaries such that the execution of $z_1$ runs from boundary $k$ to boundary $k+1$ and the execution of $z_2$ runs from boundary $k+1$ to boundary $k+2$. Thus, the composed skill terminates with history $h_{k+2}$ after $\tau_k + \tau_{k+1}$ steps. Its segment reward is $\tilde{r}_k + \gamma^{\tau_k}\, \tilde{r}_{k+1}$.

\emph{Availability.} $z_1 z_2$ is available at a history exactly when $z_1$ is. By the induction hypothesis, $z_1$ is available at $h_k$ if and only if it is available at every $h_k' \sim h_k$, so the same holds for $z_1 z_2$, which is \eqref{eq:avail}.

\emph{The two parts.} Until $h_{k+1}$, the composed skill executes $z_1$, so $(\tilde{r}_k, \tau_k, h_{k+1})$ has the same distribution given $(h_k, z_1 z_2)$ as given $(h_k, z_1)$. At $h_{k+1}$, $z_2$ starts fresh, and the distribution of the hidden state given a history does not depend on which skills produced it. So given $h_{k+1}$, the execution of $z_2$ has the same distribution as $z_2$ initiated at $h_{k+1}$. Moreover, $z_2$ is available at every observation at which $z_1$ can terminate, so if $h_{k+1}$ is in class $C'$, then $z_2 \in \mathcal{Z}_{C'}$ by \eqref{eq:avail}, and \eqref{eq:dyn-suff-gen} and \eqref{eq:rew-suff-gen} for $z_2 \in \mathcal{Z}$ give
\begin{equation}
    \mathbb{E}\big[\gamma^{\tau_{k+1}}\, \mathbf{1}\{h_{k+2} \in C''\} \mid h_{k+1}, z_1 z_2\big] = \bar{K}_\gamma(C'' \mid C', z_2),
    \qquad
    \mathbb{E}\big[\tilde{r}_{k+1} \mid h_{k+1}, z_1 z_2\big] = \bar{R}(C', z_2).
    \label{eq:second-skill}
\end{equation}

\emph{Dynamics sufficiency.} Fix a history $h_k$ in class $C$ at which $z_1 z_2$ is available, and a class $C''$. Since $h_{k+1}$ contains $h_k$ and determines $\tau_k$, the tower rule gives
\begin{align*}
    K_\gamma(C'' \mid h_k, z_1 z_2)
      &= \mathbb{E}\Big[\gamma^{\tau_k}\, \mathbb{E}\big[\gamma^{\tau_{k+1}}\, \mathbf{1}\{h_{k+2} \in C''\} \mid h_{k+1}, z_1 z_2\big] \,\Big|\, h_k, z_1\Big] \\
      &= \sum_{C'} \mathbb{E}\big[\gamma^{\tau_k}\, \mathbf{1}\{h_{k+1} \in C'\} \mid h_k, z_1\big]\, \bar{K}_\gamma(C'' \mid C', z_2)
         && \text{(By \eqref{eq:second-skill})} \\
      &= \sum_{C'} \bar{K}_\gamma(C' \mid C, z_1)\, \bar{K}_\gamma(C'' \mid C', z_2)
         && \text{(By the induction hypothesis)},
\end{align*}
which depends on $h_k$ only through $C$.

\emph{Reward sufficiency.} By linearity of expectation and the same steps,
\begin{align*}
    \mathbb{E}\big[\tilde{r}_k + \gamma^{\tau_k}\, \tilde{r}_{k+1} \mid h_k, z_1 z_2\big]
      &= \mathbb{E}[\tilde{r}_k \mid h_k, z_1] + \mathbb{E}\Big[\gamma^{\tau_k}\, \mathbb{E}\big[\tilde{r}_{k+1} \mid h_{k+1}, z_1 z_2\big] \,\Big|\, h_k, z_1\Big] \\
      &= \bar{R}(C, z_1) + \sum_{C'} \bar{K}_\gamma(C' \mid C, z_1)\, \bar{R}(C', z_2),
\end{align*}
which also depends on $h_k$ only through $C$. Hence $\sim$ is a temporally abstract bisimulation for $\mathcal{Z}^{(m+1)}$, which completes the induction.
\end{proof}

\section{Proofs from Section \ref{sec:learning}}
\label{app:proofs-elbo}

\subsection{Proof of Lemma~\ref{lem:suff-tab}}
\label{app:proof-suff-tab}

\begin{proof}
For all $h_k \sim_e h_k'$ and all $z \in \mathcal{Z}$, we have
\begin{align}
    P\big(e(h_{k+1}) \mid h_k, z\big)
      &= P\big(e(h_{k+1}) \mid h_k, e(h_k), z\big)
         && \text{(Because $e(h_k)$ is a function of $h_k$)} \\
      &= P\big(e(h_{k+1}) \mid e(h_k), z\big)
         && \text{(By \eqref{eq:dyn-suff})}.
\end{align}
We also have
\begin{align}
    P\big(e(h_{k+1}) \mid h_k', z\big)
      &= P\big(e(h_{k+1}) \mid h_k', e(h_k'), z\big)
         && \text{(Because $e(h_k')$ is a function of $h_k'$)} \\
      &= P\big(e(h_{k+1}) \mid e(h_k'), z\big)
         && \text{(By \eqref{eq:dyn-suff})} \\
      &= P\big(e(h_{k+1}) \mid e(h_k), z\big)
         && \text{(Because $e(h_k') = e(h_k)$)} \\
      &= P\big(e(h_{k+1}) \mid h_k, z\big),
\end{align}
which is the dynamics condition of the lemma. The same steps with $\tilde{r}_k$ in place of
$e(h_{k+1})$ and \eqref{eq:rew-suff} in place of \eqref{eq:dyn-suff} give the reward condition. With fixed-length skills, $\tau_k = H$. Then, for every class $C$ of $\sim_e$ we have $K_\gamma(C \mid h_k, z) = \gamma^H P\big(e(h_{k+1}) = C \mid h_k, z\big)$, so the first equality gives \eqref{eq:dyn-suff-gen}. Since $\tilde{r}_k$ has the same distribution given $(h_k, z)$ and $(h_k', z)$, it has the same expectation, which is \eqref{eq:rew-suff-gen}. Finally, every skill can be initiated at every observation, so $\mathcal{Z}(h_k) = \mathcal{Z}(h_k') = \mathcal{Z}$, which is \eqref{eq:avail}. Hence $\sim_e$ is a temporally abstract bisimulation.
\end{proof}

\subsection{Proof of Theorem~\ref{thm:elbo-tab}}
\label{app:proof-elbo-tab}

\begin{proof}

Let $P_\Phi(x, \tilde{s}, z) = p_{\mathrm{data}}(x)\, q_\Phi(\tilde{s}, z \mid x)$ be the true joint distribution of trajectories and
inferred latents. By \eqref{eq:jensen}, $\mathcal{L}(\Theta, \Phi)$ is the expectation under $P_\Phi$ of the log-ratio. Adding and
subtracting $\log p_{\mathrm{data}}(x)$ inside the expectation, the denominator becomes $P_\Phi$:
\begin{equation}
    \mathcal{L}(\Theta, \Phi) \;=\; \mathbb{E}_{P_\Phi}\Big[\log \frac{p_\Theta(x, \tilde{s}, z)}{p_{\mathrm{data}}(x)\, q_\Phi(\tilde{s}, z \mid x)}\Big]
        + \mathbb{E}_{p_{\mathrm{data}}}\big[\log p_{\mathrm{data}}(x)\big]
      \;=\; -\,\mathrm{KL}\big[P_\Phi \,\|\, p_\Theta\big] - \mathcal{H}(p_{\mathrm{data}}).
    \label{eq:J-kl}
\end{equation}
By \eqref{eq:J-kl}, the KL divergence
is nonnegative and, by realizability, attains zero, so $\mathcal{L}(\Theta, \Phi)$ is maximized if and only if
$\mathrm{KL}[P_\Phi \,\|\, p_\Theta] = 0$, i.e., $P_\Phi = p_\Theta$. Every conditional distribution of $P_\Phi$ is then the
corresponding conditional distribution of $p_\Theta$, so it suffices to compute
$p_\Theta(\tilde{r}_k, \tilde{s}_{k+1} \mid h_k, \tilde{s}_k, z_k)$ from \eqref{eq:generative}.

The generative model produces the variables in the order $\tilde{s}_0$, $o_0$, and then
$z_k, \tilde{r}_k, \tilde{s}_{k+1}, \bar{o}_k, \bar{a}_k$ for each $k$. Each factor conditions only on variables generated earlier. The history $h_k = (o_0, \bar{o}_{0:k-1}, \bar{a}_{0:k-1})$ is produced before $z_k$. We
integrate out every variable except $(h_k, \tilde{s}_k, z_k, \tilde{r}_k, \tilde{s}_{k+1})$:
\begin{itemize}
    \item The factors of the variables produced after $\tilde{s}_{k+1}$, namely
          $p_\eta(\bar{o}_k \mid \cdot)$, $\pi_\theta(a_t \mid \cdot)$ and all factors of later segments, are
          normalized distributions over their own variables. Thus, integrating out backwards, each integrates to one.
    \item The factors of the variables produced before $z_k$ involve only $h_k$, $\tilde{s}_k$ and the earlier latents and rewards. Integrating out the earlier latents and rewards leaves the model's
          marginal $p_\Theta(h_k, \tilde{s}_k)$.
    \item The factors $p_\omega(z_k \mid \tilde{s}_k)$, $p_\rho(\tilde{r}_k \mid \tilde{s}_k, z_k)$ and
          $p_\psi(\tilde{s}_{k+1} \mid \tilde{s}_k, z_k)$ only contain variables that are kept, so they remain.
\end{itemize}
Hence
\[
    p_\Theta(h_k, \tilde{s}_k, z_k, \tilde{r}_k, \tilde{s}_{k+1})
      = p_\Theta(h_k, \tilde{s}_k)\; p_\omega(z_k \mid \tilde{s}_k)\; p_\rho(\tilde{r}_k \mid \tilde{s}_k, z_k)\;
        p_\psi(\tilde{s}_{k+1} \mid \tilde{s}_k, z_k).
\]
Integrating out $\tilde{r}_k$ and $\tilde{s}_{k+1}$, the factors $p_\rho$ and $p_\psi$ are distributions over these
variables and integrate to one, so
\[
    p_\Theta(h_k, \tilde{s}_k, z_k) = p_\Theta(h_k, \tilde{s}_k)\; p_\omega(z_k \mid \tilde{s}_k).
\]
Dividing the joint by this marginal,
\begin{align*}
    p_\Theta(\tilde{r}_k, \tilde{s}_{k+1} \mid h_k, \tilde{s}_k, z_k)
      &= \frac{p_\Theta(h_k, \tilde{s}_k, z_k, \tilde{r}_k, \tilde{s}_{k+1})}{p_\Theta(h_k, \tilde{s}_k, z_k)} \\
      &= \frac{p_\Theta(h_k, \tilde{s}_k)\; p_\omega(z_k \mid \tilde{s}_k)\; p_\rho(\tilde{r}_k \mid \tilde{s}_k, z_k)\;
               p_\psi(\tilde{s}_{k+1} \mid \tilde{s}_k, z_k)}
              {p_\Theta(h_k, \tilde{s}_k)\; p_\omega(z_k \mid \tilde{s}_k)} \\
      &= p_\rho(\tilde{r}_k \mid \tilde{s}_k, z_k)\; p_\psi(\tilde{s}_{k+1} \mid \tilde{s}_k, z_k).
\end{align*}
The history $h_k$ appears only in $p_\Theta(h_k, \tilde{s}_k)$, which is common to the numerator and the
denominator and cancels, together with $p_\omega(z_k \mid \tilde{s}_k)$. Neither remaining factor depends on $h_k$.
Integrating out $\tilde{r}_k$ or $\tilde{s}_{k+1}$, and using $P_\Phi = p_\Theta$,
\[
    P_\Phi(\tilde{s}_{k+1} \mid h_k, \tilde{s}_k, z_k) = p_\psi(\tilde{s}_{k+1} \mid \tilde{s}_k, z_k),
    \qquad
    P_\Phi(\tilde{r}_k \mid h_k, \tilde{s}_k, z_k) = p_\rho(\tilde{r}_k \mid \tilde{s}_k, z_k).
\]
These identities assert that the history influences the next abstract state and the segment reward only
through $(\tilde{s}_k, z_k)$. These are the dynamics and reward sufficiency conditions
\eqref{eq:dyn-suff} and \eqref{eq:rew-suff}. Lemma~\ref{lem:suff-tab} then guarantees that the
encoder-induced partition is a temporally abstract bisimulation.
\end{proof}

\section{Implementation Details}
\label{app:arch}

\begin{table}[t]
  \centering
  \small
  \caption{Hyperparameters.}
  \label{tab:hparams}
  \begin{tabular}{lc}
    \toprule
    \textbf{Hyperparameter} & \textbf{Value} \\
    \midrule
    \multicolumn{2}{c}{\textcolor{gray}{\textit{=== Model Architecture ===}}} \\
    Abstract state dimension ($d_{\tilde{s}}$) & 32 (antmaze, humanoidmaze, cube, D4RL), 10 (pointmaze) \\
    Skill dimension ($d_z$) & 256 \\
    Hidden width & 256 \\
    Skill length ($H$) & 20 (antmaze, pointmaze), 50 (humanoidmaze), 5 (cube), \\
     & $\{5, 10, 16, 20\}$ (D4RL) \\
    Skills per window ($K$) & 4 (mazes), 8 (cube), $80/H$ (D4RL) \\
    Mean bound ($M$) & 1 \\
    State noise ($\sigma$) & 0.001 (antmaze, humanoidmaze), 0.01 (pointmaze, cube), \\
     & $\{0.001, 0.01, 0.1, 0.2, 0.3, 0.5\}$ (D4RL) \\
    \multicolumn{2}{c}{\textcolor{gray}{\textit{=== Objective Parameters ===}}} \\
    Transition KL weight & 4 \\
    Initial-state KL weight & 1 \\
    Skill KL weight & 1 \\
    Segment decoder weight & 0.1 \\
    Initial-observation decoder weight & 1 \\
    Low-level policy weight & 1 \\
    Reward head weight & 5 (antmaze, humanoidmaze, cube), 1 (pointmaze, D4RL) \\
    Standard deviation floors & 0.5 (dynamics), 0.1 (decoders), $10^{-3}$ (skill prior, policy) \\
    \multicolumn{2}{c}{\textcolor{gray}{\textit{=== Goal Relabeling (OGBench) ===}}} \\
    Boundary-state goal probability & 0.2 \\
    Future-state goal probability & 0.5 \\
    Future-state offset discount & 0.99 (0.995 humanoidmaze) \\
    Random-state goal probability & 0.3 \\
    \multicolumn{2}{c}{\textcolor{gray}{\textit{=== Training Parameters ===}}} \\
    Optimizer & Adam, no weight decay \\
    Learning rate & $5 \times 10^{-5}$ ($10^{-4}$ cube) \\
    Gradient clipping (global norm) & 10 \\
    Batch size & 100 windows of $KH{+}1$ steps \\
    Training steps & $1$M \\
    \multicolumn{2}{c}{\textcolor{gray}{\textit{=== Planner Parameters ===}}} \\
    CEM samples & 100 \\
    CEM elite ratio & 0.5 \\
    CEM iterations & 100 \\
    Planner horizon & 15 skills (OGBench), $L$ skills $=80$ steps (D4RL) \\
    Planning objective & summed segment rewards, discounted by $\gamma^{H}$ per skill \\
    Planning discount ($\gamma$) & 0.99 (0.995 humanoidmaze) \\
    \bottomrule
  \end{tabular}
\end{table}

The model comprises eight networks: the skill prior
$p_\omega$, abstract dynamics $p_\psi$, segment decoder $p_\eta$, initial-observation decoder $p_{\eta_o}$, segment score $p_\rho$, low-level policy $\pi_\theta$, the state encoder $q_\chi$, and the skill encoder $q_\phi$. We describe the conventions shared across all modules, then each network in turn. Exact hyperparameters are listed in Table~\ref{tab:hparams}.

Each model (besides the reward decoder) is a diagonal Gaussian
whose mean and standard deviation are produced by two separate heads, each a two-layer MLP consisting of a $\mathrm{Linear}$, $\mathrm{ReLU}$, and $\mathrm{Linear}$ layer. The standard deviation heads apply a Softplus. Feed-forward trunks are two-layer ReLU MLPs and all recurrent modules are GRUs.

\paragraph{Skill prior $p_\omega(z_k \mid \tilde{s}_k)$.} A feed-forward network mapping the abstract boundary state $\tilde{s}_k \in \mathbb{R}^{d_{\tilde{s}}}$ through a two-layer ReLU trunk to a diagonal Gaussian over the skill $z_k \in \mathbb{R}^{d_z}$. The prior is used during planning to provide the initial distribution over skills.

\paragraph{Abstract dynamics $p_\psi(\tilde{s}_{k+1} \mid \tilde{s}_k, z_k)$.} A feed-forward network whose input is $[\tilde{s}_k, z_k]$ and whose output is a diagonal Gaussian over the next boundary state $\tilde{s}_{k+1} \in \mathbb{R}^{d_{\tilde{s}}}$. For the \texttt{humanoidmaze} environments in OGBench, we use a GRU where $\tilde{s}_k$ is the hidden state of a GRU cell. For each dimension, the mean is a gated mix $(1 - u) * n + u * \tilde{s}_k / M$ of the current state and a tanh candidate n, so it never leaves the box its input is in ((-M, M) for the encoder's states). This is the transition model rolled forward during CEM planning.

\paragraph{Segment decoder $p_\eta(\mathcal{O}_k \mid \tilde{s}_k, z_k, \tilde{s}_{k+1}, h_k)$.} A two-layer autoregressive GRU that reconstructs the $H$ within-segment observations. The history $h_k$ is summarized by a separate single-layer GRU over the prefix observations $o_0, \dots, o_{kH}$, and its output at the
boundary is concatenated with $[\tilde{s}_k, z_k, \tilde{s}_{k+1}]$. This vector initializes the GRU's hidden state and is input at every step. We use teacher forcing such that the step that predicts $o_t$ also receives an embedding of the previous ground-truth observation $o_{t-1}$, starting from the boundary
observation $o_{kH}$. Each step emits a diagonal Gaussian over $o_t$. 

\paragraph{Initial-observation decoder $p_{\eta_o}(o_0 \mid \tilde{s}_0)$.} A feed-forward network mapping the initial abstract state $\tilde{s}_0$ to a diagonal Gaussian over the first observation $o_0$. 

\paragraph{Reward decoder $p_\rho(\tilde{r}_k \mid \tilde{s}_k, z_k, g)$.} A feed-forward network with three hidden ReLU layers mapping $[\tilde{s}_k, z_k, g]$ to the logit of a Bernoulli distribution over the segment reward, which is $1$ if the goal is reached within segment $k$ and $0$ otherwise. For the locomotion environments, the goal $g$ is the target $xy$ position in the mazes. For the manipulation cube environment, the goal is the target cube position.

\paragraph{Low-level policy $\pi_\theta(a_t \mid o_t, z_k)$.} A feed-forward network mapping the low-level observation $o_t$ and the skill $z_k$ to a diagonal Gaussian over the low-level action $a_t \in \mathbb{R}^{d_a}$. o

\paragraph{State encoder $q_\chi(\tilde{s}_k \mid h_k)$.} A four-layer \emph{unidirectional} GRU over the prefix history. Each observation is first embedded by a two-layer MLP (with ReLU activations after each) and concatenated with the previous action before entering the GRU. The hidden states read off at the segment boundaries (indices $kH$ for $k=0, \dots, K$) give the $K{+}1$ boundary posteriors. The mean head maps each to
$M\tanh(\mu/M)$, and the standard deviation is the fixed constant $\sigma$.

\paragraph{Skill posterior $q_\phi(z_k \mid \mathcal{O}_k, \mathcal{A}_k, \tilde{s}_k)$.} A four-layer \emph{bidirectional} GRU over $o_{kH+1}, \dots, o_{(k+1)H}$. Each observation is embedded with a two-layer MLP (with ReLU activations after each) and concatenated with the prior action as GRU input. The GRU's output at the last step of the segment from both directions is concatenated with an embedding of $\tilde{s}_k$ and passed through the mean/std heads to give a diagonal Gaussian over $z_k$. 

\section{Environments}

\textbf{Locomotion tasks from D4RL} \cite{fu2020d4rl}:  
In Maze2D, a point mass moves through a 2D maze. In AntMaze, an 8-DoF quadruped ``ant'' robot has to
locomote through the same layout. Although the two domains share the same navigation problem, they differ in their low-level control. In comparison to Maze2D, most of AntMaze states describe the robot's body rather than where it is in the maze.

For AntMaze we use the \textit{diverse} dataset (\texttt{antmaze-medium-diverse}), in which the ant is commanded to random goals from random start locations. For Maze2D we use \texttt{maze2d-medium}, which was collected by a planner steering the point mass to randomly sampled waypoints. In AntMaze, the agent must navigate from a fixed start in one corner of the maze to a goal in the opposite corner. In Maze2D, following \citet{freed2023oposm}, we resample the goal uniformly over the cells of the maze at every trial, so the agent must reach arbitrary goals rather than a single fixed one.

\textbf{Locomotion and manipulation tasks from OGBench} \cite{park2025ogbench}:
\textit{Pointmaze} controls a 2-DoF point mass, \textit{Antmaze} controls an 8-DoF quadruped, and \textit{Humanoidmaze} controls a 21-DoF humanoid. In all three maze domains the goal is a target $xy$ location. In \textit{cube-single}, a 6-DoF  UR5e arm with a parallel Robotiq 2F-85 gripper has to move a cube to a target pose.

We follow OGBench's evaluation protocol. Each environment defines five evaluation tasks representing five unique start-goal pairs. Following \cite{park2025ogbench}, we report success averaged over the five tasks. An episode succeeds if the agent reaches the goal within the environment's threshold ($0.5$ for antmaze and humanoidmaze, $1.0$ for pointmaze, $0.04$\,m between the cube and its target for cube-single). Episodes end at the goal or at the environment's step limit (1000 steps for pointmaze and antmaze, 2000 for humanoidmaze, 200 for cube-single).

\section{Additional Results and Visualizations}
\label{app:addresults}

\begin{figure}[t]
  \centering
  \includegraphics[width=\linewidth]{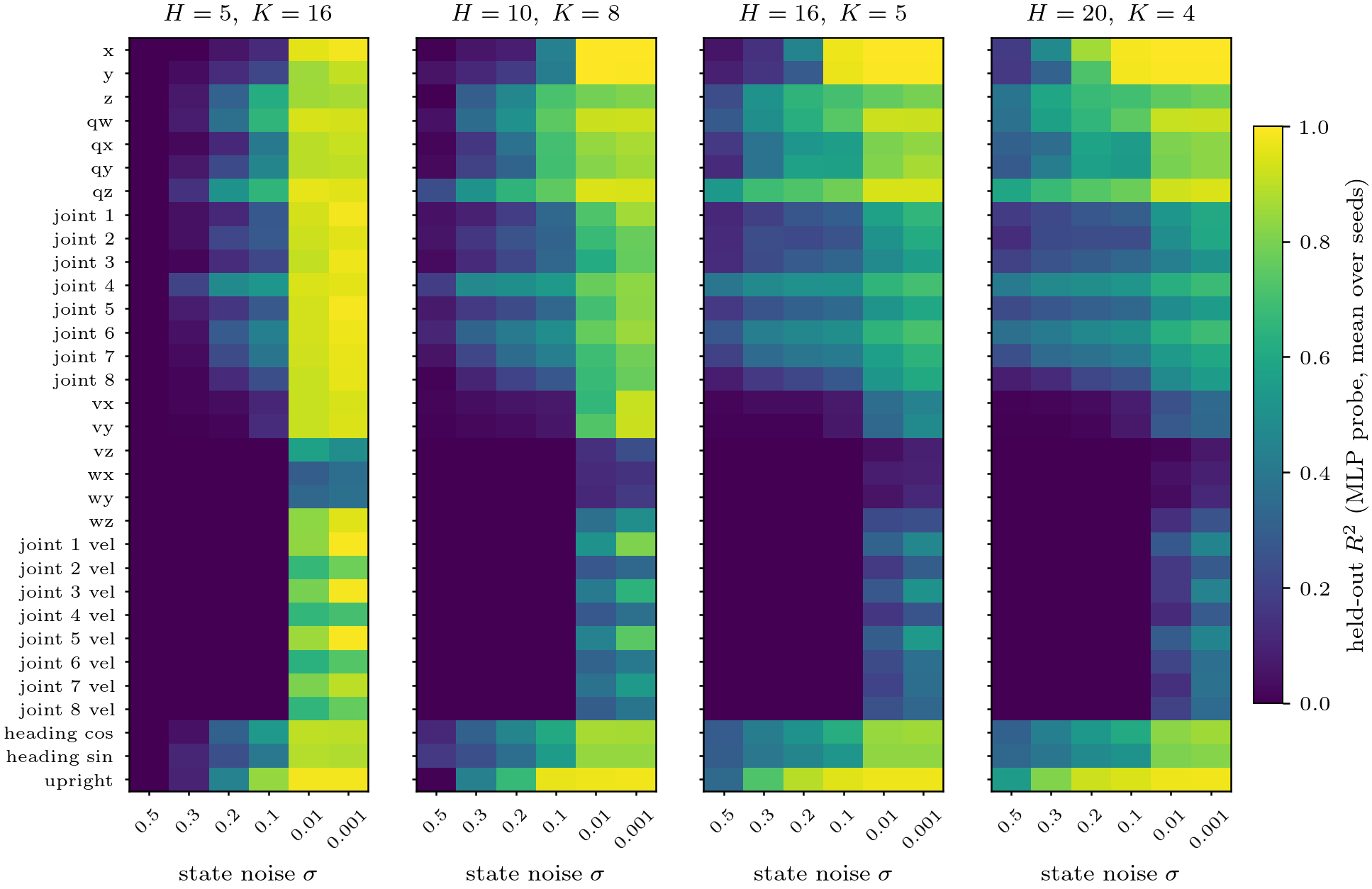}
  \caption{Held-out $R^2$ of each observation component against different values of $\sigma$ for all $H$. Each square is $R^2$ averaged over $3$ seeds. As $\sigma$ increases, longer skills put more of the abstract state into position and less into the joint angles and velocities.}
  \label{fig:summary-components}
\end{figure}

Unlike the previous experiment (Figure~\ref{fig:content-bits}) where the MLP probe is trained on groups of the observation components, here, we train an MLP probe on each observation component separately. Figure \ref{fig:summary-components} shows similar patterns, where longer skills encode more positional information even with a tighter information cap. Each of the $8$ joint angles and each of the $8$ joint velocities is harder to decode for longer skills.

\begin{figure}[t]
  \centering
  \begin{subfigure}[t]{\linewidth}
    \centering
    \includegraphics[width=\linewidth]{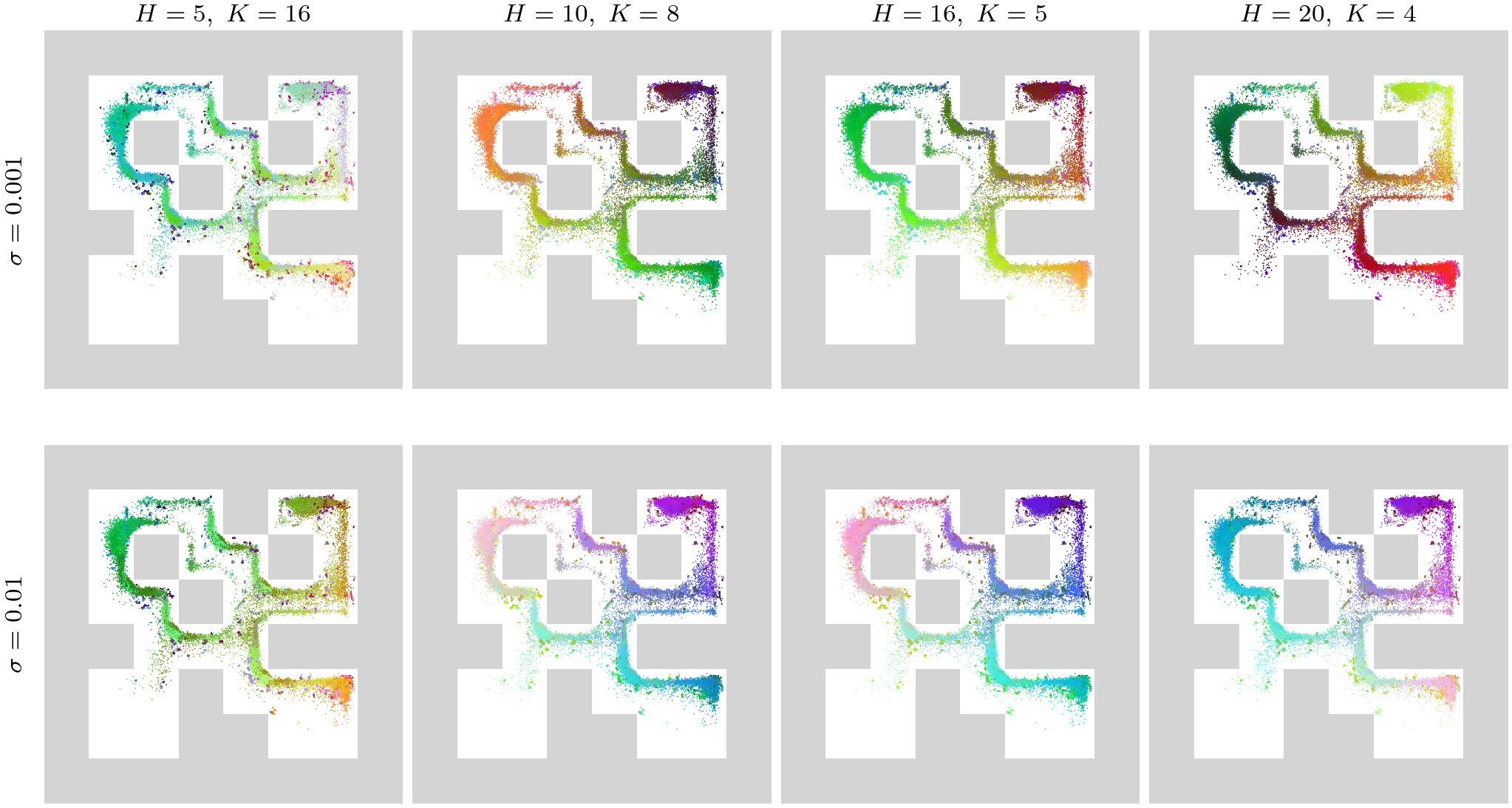}
    \caption{First three principal components of $\mu(h_k)$ as RGB.}
    \label{fig:content-pca}
  \end{subfigure}
  \begin{subfigure}[t]{\linewidth}
    \centering
    \includegraphics[width=\linewidth]{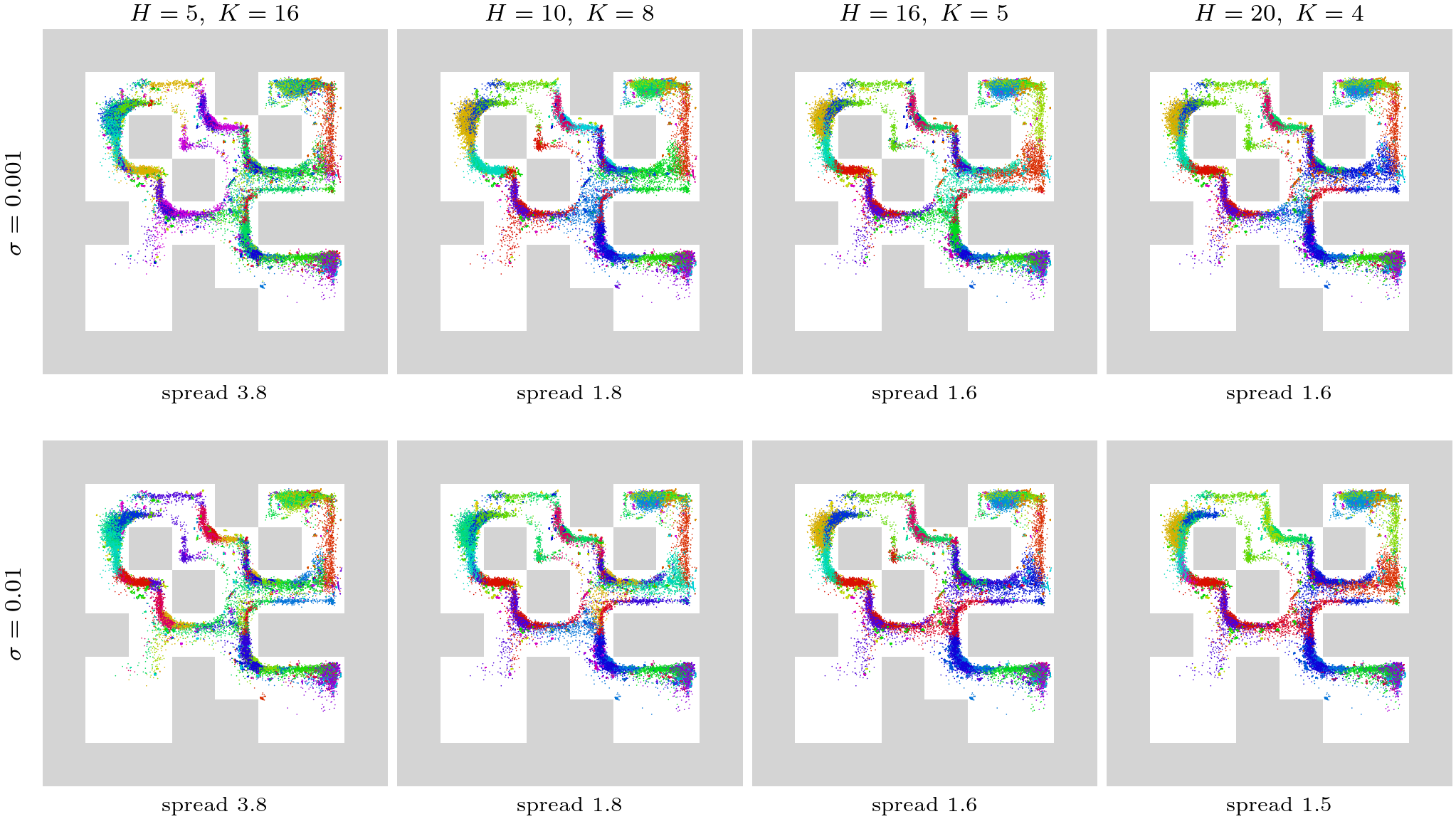}
    \caption{k-means clusters of $\mu(h_k)$ ($40$ clusters), with the cluster spread (mean over $3$ seeds) below each panel.}
    \label{fig:content-clusters}
  \end{subfigure}
  \caption{Dataset states plotted at their true position in AntMaze-Medium and coloured by their abstract state, for four
  skill lengths and the two values of $\sigma$ at which planning has not collapsed ($60{,}000$ states, seed $0$).
  For longer skills, nearby states have similar abstract states and clusters cover compact stretches of the maze.}
  \label{fig:content-maps}
\end{figure}

In Figure~\ref{fig:content-bits} and Figure~\ref{fig:summary-components}, we showed with MLP probes that longer skills encouraged the abstract state to expend bits on encoding positional information. Now, we show this claim qualitatively. Figure~\ref{fig:content-maps} draws $60{,}000$ dataset states at their true position, colored by their abstract state. We choose $\sigma = 0.001$ and $0.01$ because planning has not collapsed at these values. In Figure~\ref{fig:content-pca} the plotted color is the first three principal components of the posterior mean $\mu(h_k)$ as RGB. For $H = 5$, the map is speckled, where nearby states often take different colors. This shows that the abstract state also varies substantially with the body state (non-positional information). For $H \ge 10$, the color changes more smoothly along the corridors, signifying that the abstract states contain more positional information relative to the other groups.

In Figure~\ref{fig:content-clusters} the color is a k-means cluster of $\mu(h_k)$. Qualitatively, for long skills the clusters cover compact stretches of the maze, while for $H = 5$ they are spread out and overlap. We measure this quantitatively with the \emph{cluster spread}: for each cluster, $\sqrt{\operatorname{Var}(x) + \operatorname{Var}(y)}$ of its states, averaged
over clusters with weights proportional to their size. At $\sigma = 0.001$ the spread is $3.8$ for $H = 5$ against $1.8$, $1.6$ and $1.6$ for
$H = 10, 16, 20$. 

\end{document}